\documentclass{article} 
\usepackage[preprint]{colm2026_conference}

\usepackage{microtype}
\usepackage{hyperref}
\usepackage{url}
\usepackage{graphicx}
\usepackage{listings}
\usepackage{xcolor}
\usepackage{xspace}
\usepackage{wrapfig}
\usepackage{booktabs}
\usepackage{multirow}
\usepackage{algorithm}
\usepackage{algpseudocode}
\usepackage{tcolorbox}
\tcbuselibrary{breakable,skins}
\tcbuselibrary{breakable}
\usepackage{soul}
\usepackage{subcaption}
\usepackage{amsmath}
\usepackage{fancyvrb}
\usepackage{enumitem}
\usepackage{fvextra}

\definecolor{promptbg}{RGB}{248,249,250}
\definecolor{promptframe}{RGB}{140,140,140}
\definecolor{plcolor}{RGB}{0,80,180}

\newtcolorbox{promptbox}[1]{
  breakable,
  colback=promptbg,
  colframe=promptframe,
  fonttitle=\small\bfseries\sffamily,
  title={#1},
  left=4pt, right=4pt, top=4pt, bottom=4pt
}

\usepackage{listings}
\usepackage{xcolor}
\newcommand{\ph}[1]{\textcolor{plcolor}{\textit{$\langle$#1$\rangle$}}}

\usepackage{lineno}
\newcommand{\deontic}{\textsc{DeonticBench}\xspace}

\definecolor{darkblue}{rgb}{0, 0, 0.5}
\hypersetup{colorlinks=true, citecolor=darkblue, linkcolor=darkblue, urlcolor=darkblue}

\title{\deontic: A Benchmark for Reasoning over Rules}

\newcommand{\aspace}{\hspace{1em}}
\newcommand{\jhu}{$^{\heartsuit}$}
\newcommand{\telecom}{$^{\spadesuit}$}

\author{%
    \textbf{Guangyao Dou}\jhu\thanks{Correspondence to: \texttt{gdou1@jhu.edu}.}\aspace
    \textbf{Luis Brena}\jhu\aspace
    \textbf{Akhil Deo}\jhu\aspace
    \textbf{William Jurayj}\jhu\aspace
    \textbf{Jingyu Zhang}\jhu\aspace\\[0.4ex]
    \textbf{Nils Holzenberger}\telecom\aspace
    \textbf{Benjamin Van Durme}\jhu\aspace\\[0.4ex]
    \jhu Johns Hopkins University\aspace\telecom T\'el\'ecom Paris, Institut Polytechnique de Paris\\[0.4ex]
}

\begin{document}

\ifcolmsubmission
\linenumbers
\fi

\maketitle

\begin{abstract}
Reasoning with complex, context-specific rules remains challenging for large language models (LLMs). In legal and policy settings, this manifests as \emph{deontic reasoning}: reasoning about obligations, permissions, and prohibitions under explicit rules. While many recent benchmarks emphasize short-context mathematical reasoning, fewer focus on long-context, high-stakes deontic reasoning. To address this gap, we introduce \deontic, a benchmark of 6,232 tasks across U.S. federal taxes, airline baggage policies, U.S. immigration administration, and U.S. state housing law. These tasks can be approached in multiple ways, including direct reasoning in language or with the aid of symbolic computation. Besides free-form chain-of-thought reasoning, \deontic enables an optional solver-based workflow in which models translate statutes and case facts into executable Prolog, leading to formal problem interpretations and an explicit program trace. We release reference Prolog programs for all instances. Across frontier LLMs and coding models, best hard-subset performance reaches only 44.4\% on SARA Numeric and 46.6 macro-F1 on Housing. We further study training with supervised fine-tuning and reinforcement learning for symbolic program generation. Although training improves Prolog generation quality, current RL methods still fail to solve these tasks reliably. Overall, \deontic provides a benchmark for studying context-grounded rule reasoning in real-world domains under both symbolic and non-symbolic settings.\footnote{Code and data are available at \href{https://github.com/guangyaodou/DeonticBench}{guangyaodou/DeonticBench}.}
\end{abstract}

\section{Introduction}
In February 2026, the New York State Senate Committee passed Senate Bill S7263 \citep{nysenate_s7263_2026}, establishing liability for harms caused by AI chatbots impersonating licensed professionals. This case reflects a broader trend: as large language models (LLMs) are deployed in high-stakes settings such as healthcare \citep{he2025survey} and legal systems \citep{dehghani2025large}, errors can create legal, financial, and safety risks. Yet even with rapid progress and wider adoption, LLMs remain prone to hallucinations \citep{huang2025survey, bengio2026international}. Ensuring that LLMs behave faithfully, especially when reasoning over formal rules, is therefore critical.

\begin{figure}[ht]
    \centering
    \includegraphics[width=1\textwidth]{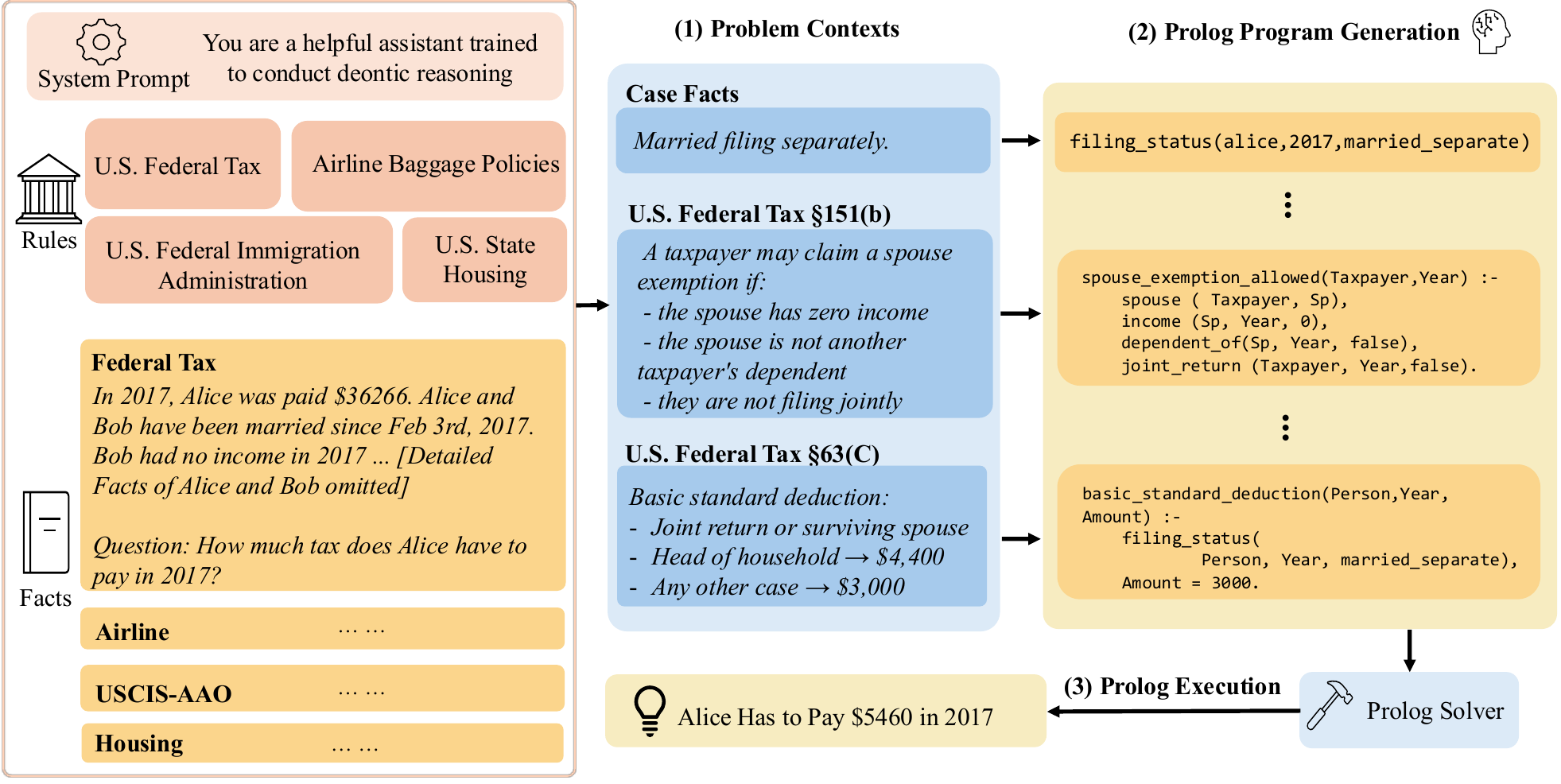} 
     \caption{Walkthrough of a \deontic instance in the symbolic setting. (1) Given the full problem context, the model performs deontic reasoning to identify and apply the relevant rules. (2) The LLM translates the problem into Prolog code. (3) The generated Prolog is executed by SWI-Prolog solver. The illustrated example is a 2017 tax-liability case.}
\label{fig:introduction}
\end{figure}

Language model reasoning using chain-of-Thought \citep[CoT;][]{wei2022chain, guo2025deepseek} shows strong gains on challenging tasks, yet exhibits key failure modes, including harder-to-detect hallucinations and errors on simple cases \citep{cheng2025chain, shojaee2025illusion}. In high-stakes settings, these limitations are especially concerning because models must be not only accurate but also faithful to explicit rules, making \emph{deontic reasoning}---reasoning about obligations, permissions, and prohibitions under formal statutes and policies---an important capability to evaluate. In parallel, symbolic reasoning maps natural-language problems into executable code run by symbolic solvers, offering a potential path toward more faithful and interpretable rule-grounded reasoning \citep{pan2023logic, mahdavi2024leveraging, ye2023satlm, lyu2023faithful}.

To study this capability, we introduce \deontic, a benchmark of 6,232 tasks spanning U.S. federal taxes, airline baggage policies, U.S. immigration administration, and U.S. state housing law. \deontic extends the mathematical-reasoning and STEM-benchmark paradigm \citep{cobbe2021training, glazer2024frontiermath, hendrycks2020measuring} to high-stakes, statute-grounded rule reasoning. It includes both quantitative tasks and binary deontic-logic tasks. Building on prior work on statutory reasoning with symbolic solvers \citep{holzenberger-van-durme-2023-connecting, jurayj2026language}, we include a solver-assisted workflow in which models can map rules and case facts to executable Prolog code \citep{colmerauer1973systeme}. This workflow provides auditable reasoning artifacts---a formal interpretation of the input and an explicit program trace---while our evaluation also includes non-symbolic settings. We release reference Prolog code for all instances as artifacts for analysis and training, and we evaluate models under both symbolic and non-symbolic settings. Previous work suggests that the advantage of the symbolic setting is task-dependent---some tasks benefit more from symbolic methods, while others do not \citep{chen2024steering}.

We evaluate eight general-purpose LLMs (including five reasoning models) and three coding models on \deontic. We find that all models struggle on the hard subsets of the benchmark (\S~\ref{sec:hard-set-construction}), and that increased reasoning effort does not consistently improve performance. We further post-train Qwen2.5-32B-Instruct \citep{qwen2.5} with supervised fine-tuning and both on-policy and off-policy reinforcement learning, improving performance, but yielding limited gains in end-to-end task accuracy.

In summary, our contributions are threefold: (1)~we introduce \deontic, the largest benchmark to date for executable rule-based reasoning in real-world, high-stakes domains; (2)~we provide a comprehensive evaluation and analysis of failure modes in current frontier models, finding that all evaluated models struggle on hard subsets; and (3)~we investigate training strategies, highlighting the limitations of existing reinforcement learning approaches and the need for further advances.

\section{\deontic Dataset Construction}
\label{dataset_construction_overall}
               
\deontic is a multi-domain deontic reasoning benchmark that enables symbolic and non-symbolic settings. One symbolic workflow is illustrated in Figure \ref{fig:introduction}.

\subsection{Benchmark Construction}
\label{dataset_construction}

\deontic covers four real-world rule domains: U.S. federal taxes (SARA) \citep{holzenberger-van-durme-2023-connecting}, airline baggage policies (Airline) \citep{zhou2025rulearena}, U.S. state housing law (Housing) \citep{zheng2025reasoning}, and U.S. immigration administration (USCIS-AAO). All tasks follow a unified input-output format, but the domains vary in answer type and reasoning demands, including numeric computation, logical entailment, and multi-factor legal decision-making.

The benchmark expands prior data rather than simply reusing it. We combine three existing domains (SARA, Airline, and Housing) with a newly constructed USCIS-AAO dataset, and convert all domains to a unified task format. We also add reference Prolog programs for each instance to encode corresponding rules and case facts. This supports consistent cross-domain evaluation in both symbolic and non-symbolic settings. Representative examples from each domain are provided in Appendix~\ref{appendix:examples}.

\subsection{Curated and Expanded Source Datasets}

\paragraph{SARA (U.S. Federal Tax).}
The SARA subset in \deontic adapts the original SARA \citep{holzenberger-van-durme-2023-connecting} and retains two task types. SARA Numeric asks models to compute federal tax liability from income and filing status. SARA Binary asks whether a statute-grounded claim follows from the statute and case facts (e.g., ``Entailment or Contradiction? Alice's total exemption for 2015 under Section 151(a) is equal to \$4000.'').

The original SARA dataset provides Prolog code for each case, but assumes the full tax statute has already been encoded as background rules. In contrast, \deontic adopts a more realistic  setup: for each case, the model requires Prolog generation on the fly from the statute and case context, without relying on a pre-encoded background-rule program.

\paragraph{Airline (Airline Baggage Policies).}
The Airline cases are a subset from RuleArena \citep{zhou2025rulearena}. The Airline domain focuses on numeric baggage fee computation. Given a passenger's ticket status, baggage attributes, and the corresponding airline policy (statute), models should determine the total applicable fees by correctly applying structured airline baggage policy rules. 

\paragraph{Housing (U.S. State Housing).}
Housing cases in \deontic are the Housing Statute QA problems from a legal retrieval setting using RAG \citep{zheng2025reasoning}, which poses binary yes/no questions grounded in state-level housing statutes. Specifically, we look at the first split (rc\_questions) that contains 6,853 question-answer pair examples with labeled supporting statutes. In these cases, models must identify the relevant statutory provisions and apply them to case-specific scenarios across diverse jurisdictions. 

\subsection{New Dataset: USCIS-AAO}

USCIS-AAO is a binary immigration-appeal dataset in which each case is labeled as accepted or dismissed. Solving each task requires applying multifactor legal criteria from Administrative Appeals Office case records.

\paragraph{Data Source and Sampling}

The dataset was constructed from publicly available Administrative Appeals Office (AAO) non-precedent decisions.\footnote{\href{https://www.uscis.gov/administrative-appeals/aao-decisions/aao-non-precedent-decisions}{USCIS AAO non-precedent decisions}.} These records document appeals of prior United States Citizenship and Immigration Services (USCIS) decisions and U.S. Immigration and Customs Enforcement (ICE) determinations. We retained only cases with a consistent structure suitable for dataset construction and analysis. Specifically, a case was considered valid if it contained three distinct, non-overlapping sections: \textit{Law} (statutory basis), \textit{Analysis} (case reasoning), and \textit{Order} (final decision).

Cases were collected in USCIS upload order, yielding 6,483 valid cases from 2022--2025. We restricted the study to this time window to keep analyses within a relatively consistent precedent regime: newer decisions can complement, modify, or reinforce existing interpretations (Appendix \ref{appendix:uscis-dataset-process}).

We then selected 242 cases using stratified sampling by decision label and year. The source distribution was highly imbalanced (Appendix~\ref{appendix:uscis-dataset-stats}), with only $20.1\%$ of valid cases labeled \textit{Accepted}. This imbalance constrained the size of the final balanced subset. After filtering and balancing, the final dataset contains 242 cases: 121 \textit{Accepted} and 121 \textit{Dismissed}.

\paragraph{Label Mapping and Fact Extraction}

Within this framework, accepted decisions refer to cases that are either remanded for further revision or where the benefit petition is directly approved. In this study, remanded cases were classified as accepted. A remand indicates that the appeal has been accepted, although a final decision regarding the benefit cannot yet be rendered, typically due to missing evidence or procedural errors by the original office (Appendix \ref{appendix:uscis-dataset-process}). More detailed dataset statistics can be found in Table \ref{tab:uscis-dataset-stats}.

A primary challenge in the construction process is that inferring the decision (ground truth) of each case from the analysis section is not valid, as this section contains AAO judgments, assessments, and strong indications of the final decision. To address this issue, a facts-only narrative was generated and stored as a new field within the artifact. This extraction was conducted using the GPT-5-mini model, followed by human revision. The prompt utilized is provided in appendix \ref{appendix:additional-prompts}. The released artifact presents each case divided into three raw sections and includes the model-generated facts-only narrative.

\setlength{\intextsep}{0.4em}
\setlength{\columnsep}{0.8em}
\begin{wraptable}{r}{0.54\textwidth}
\vspace{-0.5em}
\centering
\footnotesize
\setlength{\tabcolsep}{3.2pt}
\renewcommand{\arraystretch}{1.1}

\caption{Statistics of datasets, including counts of tasks and average token lengths (via \texttt{cl100k\_base} tokenizer) by split.}
\label{tab:data_statistics}

\begin{tabular}{llrrrr}
\toprule
\multirow{2}{*}{Domain} & \multirow{2}{*}{Split} & \multirow{2}{*}{\#Tasks} & \multicolumn{3}{c}{Tokens (Mean)} \\
\cmidrule(lr){4-6}
& & & Statute & Case & Prolog \\
\midrule

\multirow{2}{*}{SARA Numeric}
& Whole & 100 & 6118 & 83 & 945 \\
& Hard  & 35  & 6118 & 89 & 1236 \\
\midrule
\multirow{2}{*}{SARA Binary}
& Whole & 276 & 6118 & 47 & 361 \\
& Hard  & 30  & 6118 & 52 & 453 \\
\midrule
\multirow{2}{*}{Airline}
& Whole & 300 & 3626 & 187 & 880 \\
& Hard  & 80  & 3626 & 197 & 1034 \\
\midrule
\multirow{2}{*}{Housing}
& Whole & 5314 & 2219 & 23 & 1350 \\
& Hard  & 78   & 588  & 23 & 680 \\
\midrule
\multirow{2}{*}{USCIS-AAO}
& Whole & 242 & 436 & 396 & 884 \\
& Hard  & 28  & 369 & 408 & 956 \\

\bottomrule
\end{tabular}
\vspace{-0.5em}
\end{wraptable}

\subsection{Prolog Generation Pipeline}
We consider two LLM-to-Prolog pipelines. One pre-translates each unique statute into a reusable Prolog library. The other generates statute modules on the fly per instance and compile-checks them in SWI-Prolog solver \citep{wielemaker:2011:tplp}.

We adopt the second design because failures are localized to each statute-question pair, and successful generations transfer cleanly to updated corpora without maintaining a global library.

Each generated artifact (statute modules + question program) is executed and counted as passing only if it compiles/runs cleanly and its final answer matches the gold label. To increase usable artifacts, we allow up to two attempts per instance: on failure, we re-prompt with the prior code and SWI-Prolog feedback; if the second attempt fails, we discard the instance.

We additionally performed human validation on sampled generations to curate few-shot demonstrations used for subsequent generation rounds. For USCIS-AAO, we applied extra prompt constraints to reduce syntactic errors. Detailed prompts, failure categories, curation workflow, and query templates are provided in Appendix~\ref{appendix:prolog-generation-details}.

\subsection{Hard Set Construction}
\label{sec:hard-set-construction}
We believe the hard split of \deontic is the most enabling contribution of this work. As frontier models increasingly saturate many existing benchmarks, aggregate scores over thousands of instances can obscure meaningful differences while remaining expensive to compute. Therefore, we carefully curate a small subset of problems across the domains that allows for more focused and affordable experimentation.

We construct hard subsets through a two-stage process: automated filtering followed by iterative human validation. We then split validated hard instances into a held-out \emph{hard set} for evaluation and the remaining portion returned to the main training pool. Full construction details are provided in Appendix~\ref{appendix:additional-hard-set-construction}.

\subsection{Dataset Statistics}

Table~\ref{tab:data_statistics} summarizes the size and token statistics of \deontic across domains and splits. In SARA and Airline, the statutes (e.g., tax law or carrier policies) are fixed across cases. On the other hand, Housing and USCIS involve case-specific statutes or legal references, leading to shorter but more variable inputs. We can also observe that the hard subsets generally require more complex reasoning, and hence, longer Prolog codes to solve. Concretely, mean Prolog length increases in 4 out of 5 domains: 945 to 1236 (SARA Numeric), 361 to 453 (SARA Binary), 880 to 1034 (Airline), and 884 to 956 (USCIS-AAO).

\begin{table*}[t]
  \centering
  \footnotesize
  \renewcommand{\arraystretch}{1.15}
  \setlength{\tabcolsep}{5pt}
  \begin{tabular}{@{}ll ccccc@{}}
    \toprule
    & & \multicolumn{2}{c}{\textbf{Accuracy}} & \multicolumn{3}{c}{\textbf{Macro F1}} \\
    \cmidrule(lr){3-4} \cmidrule(lr){5-7}
    \textbf{Model} & \textbf{Setting}
      & \textbf{SARA Num.} & \textbf{Airline}
      & \textbf{SARA Bin.} & \textbf{USCIS-AAO} & \textbf{Housing} \\
    \midrule
    \multirow{3}{*}{GPT-4.1} & Few-Shot & $23.7{\scriptstyle^{+13.5}_{-12.3}}$ & $41.5{\scriptstyle^{+11.0}_{-11.5}}$ & $39.1{\scriptstyle^{+17.2}_{-18.1}}$ & $53.0{\scriptstyle^{+18.3}_{-18.6}}$ & $46.6{\scriptstyle^{+11.6}_{-10.8}}$ \\
     & Zero-Shot & $6.7{\scriptstyle^{+10.5}_{-6.7}}$ & $1.7{\scriptstyle^{+3.3}_{-1.7}}$ & $40.5{\scriptstyle^{+16.4}_{-18.0}}$ & $55.5{\scriptstyle^{+18.7}_{-19.1}}$ & $44.7{\scriptstyle^{+10.5}_{-10.9}}$ \\
     & Direct & $18.8{\scriptstyle^{+12.6}_{-10.2}}$ & $6.7{\scriptstyle^{+5.8}_{-5.4}}$ & $30.3{\scriptstyle^{+15.5}_{-15.2}}$ & $50.9{\scriptstyle^{+16.5}_{-19.6}}$ & $20.2{\scriptstyle^{+10.1}_{-7.9}}$ \\
    \midrule
    \multirow{3}{*}{O3} & Few-Shot & $15.2{\scriptstyle^{+13.3}_{-12.4}}$ & $90.8{\scriptstyle^{+5.5}_{-7.0}}$ & $29.5{\scriptstyle^{+16.9}_{-15.6}}$ & $49.4{\scriptstyle^{+18.4}_{-18.1}}$ & $43.0{\scriptstyle^{+10.0}_{-10.1}}$ \\
     & Zero-Shot & $44.4{\scriptstyle^{+15.6}_{-15.8}}$ & $18.5{\scriptstyle^{+9.0}_{-8.5}}$ & $32.3{\scriptstyle^{+17.5}_{-15.4}}$ & $48.5{\scriptstyle^{+18.7}_{-18.5}}$ & $39.6{\scriptstyle^{+11.6}_{-10.7}}$ \\
     & Direct & $33.5{\scriptstyle^{+15.1}_{-13.5}}$ & $37.8{\scriptstyle^{+10.9}_{-11.5}}$ & $59.5{\scriptstyle^{+18.6}_{-16.9}}$ & $52.6{\scriptstyle^{+19.5}_{-17.7}}$ & $20.8{\scriptstyle^{+9.6}_{-8.4}}$ \\
    \midrule
    \multirow{3}{*}{GPT-5.1} & Few-Shot & $33.1{\scriptstyle^{+15.5}_{-15.9}}$ & $52.3{\scriptstyle^{+11.4}_{-11.1}}$ & $28.7{\scriptstyle^{+17.7}_{-14.8}}$ & $61.4{\scriptstyle^{+17.2}_{-18.5}}$ & $46.8{\scriptstyle^{+11.1}_{-10.8}}$ \\
     & Zero-Shot & $44.0{\scriptstyle^{+15.9}_{-15.5}}$ & $40.2{\scriptstyle^{+11.1}_{-11.4}}$ & $20.2{\scriptstyle^{+15.2}_{-13.5}}$ & $65.9{\scriptstyle^{+17.8}_{-18.3}}$ & $41.2{\scriptstyle^{+11.1}_{-10.9}}$ \\
     & Direct & $15.9{\scriptstyle^{+12.7}_{-10.2}}$ & $28.4{\scriptstyle^{+9.1}_{-9.6}}$ & $54.0{\scriptstyle^{+18.1}_{-18.8}}$ & $71.5{\scriptstyle^{+16.6}_{-16.4}}$ & $18.4{\scriptstyle^{+8.3}_{-7.6}}$ \\
    \midrule
    \multirow{3}{*}{GPT-5.2} & Few-Shot & $17.1{\scriptstyle^{+11.4}_{-11.4}}$ & $77.9{\scriptstyle^{+8.4}_{-9.1}}$ & $24.3{\scriptstyle^{+18.2}_{-14.7}}$ & $37.6{\scriptstyle^{+19.3}_{-16.6}}$ & $41.1{\scriptstyle^{+10.0}_{-9.9}}$ \\
     & Zero-Shot & $27.6{\scriptstyle^{+15.2}_{-13.3}}$ & $2.5{\scriptstyle^{+3.7}_{-2.5}}$ & $16.0{\scriptstyle^{+13.9}_{-12.6}}$ & $51.5{\scriptstyle^{+19.3}_{-17.1}}$ & $33.4{\scriptstyle^{+10.8}_{-10.3}}$ \\
     & Direct & $20.7{\scriptstyle^{+13.6}_{-12.1}}$ & $29.9{\scriptstyle^{+10.1}_{-9.9}}$ & $25.8{\scriptstyle^{+17.1}_{-15.2}}$ & $58.3{\scriptstyle^{+18.9}_{-18.2}}$ & $17.4{\scriptstyle^{+8.4}_{-7.7}}$ \\
    \midrule
    \multirow{3}{*}{Kimi~K2 Instruct} & Few-Shot & $9.3{\scriptstyle^{+10.7}_{-6.5}}$ & $42.6{\scriptstyle^{+9.9}_{-11.4}}$ & $48.5{\scriptstyle^{+19.0}_{-18.6}}$ & $38.2{\scriptstyle^{+16.9}_{-13.9}}$ & $37.1{\scriptstyle^{+10.9}_{-9.3}}$ \\
     & Zero-Shot & $10.1{\scriptstyle^{+12.8}_{-7.2}}$ & $0.0{\scriptstyle^{+0.0}_{-0.0}}$ & $52.7{\scriptstyle^{+18.2}_{-18.3}}$ & $43.4{\scriptstyle^{+17.7}_{-17.1}}$ & $39.8{\scriptstyle^{+11.4}_{-11.3}}$ \\
     & Direct & $8.4{\scriptstyle^{+8.7}_{-8.4}}$ & $0.9{\scriptstyle^{+2.9}_{-0.9}}$ & $68.4{\scriptstyle^{+17.0}_{-17.6}}$ & $51.8{\scriptstyle^{+18.6}_{-17.3}}$ & $24.9{\scriptstyle^{+9.7}_{-8.9}}$ \\
    \midrule
    \multirow{3}{*}{Claude~Sonnet~4.5} & Few-Shot & $21.6{\scriptstyle^{+12.7}_{-13.0}}$ & $85.8{\scriptstyle^{+6.7}_{-8.3}}$ & $28.7{\scriptstyle^{+16.9}_{-14.8}}$ & $63.1{\scriptstyle^{+17.0}_{-17.3}}$ & $42.9{\scriptstyle^{+11.0}_{-11.7}}$ \\
     & Zero-Shot & $21.9{\scriptstyle^{+15.2}_{-13.4}}$ & $5.7{\scriptstyle^{+5.6}_{-4.4}}$ & $43.9{\scriptstyle^{+16.8}_{-16.8}}$ & $63.0{\scriptstyle^{+18.5}_{-17.9}}$ & $45.0{\scriptstyle^{+11.5}_{-11.3}}$ \\
     & Direct & $41.2{\scriptstyle^{+15.9}_{-15.5}}$ & $6.2{\scriptstyle^{+5.1}_{-4.9}}$ & $70.0{\scriptstyle^{+15.6}_{-17.0}}$ & $7.2{\scriptstyle^{+8.0}_{-7.2}}$ & $32.1{\scriptstyle^{+9.4}_{-8.1}}$ \\
    \midrule
    \multirow{3}{*}{Gemini~2.5~Flash} & Few-Shot & $2.7{\scriptstyle^{+5.9}_{-2.7}}$ & $28.4{\scriptstyle^{+10.4}_{-9.6}}$ & $44.7{\scriptstyle^{+19.4}_{-17.6}}$ & $31.8{\scriptstyle^{+15.8}_{-14.1}}$ & $43.4{\scriptstyle^{+11.5}_{-10.6}}$ \\
     & Zero-Shot & $0.9{\scriptstyle^{+4.8}_{-0.9}}$ & $0.6{\scriptstyle^{+1.9}_{-0.6}}$ & $16.5{\scriptstyle^{+15.5}_{-13.1}}$ & $33.4{\scriptstyle^{+17.1}_{-13.4}}$ & $46.0{\scriptstyle^{+11.2}_{-11.0}}$ \\
     & Direct & $30.5{\scriptstyle^{+15.2}_{-13.3}}$ & $18.1{\scriptstyle^{+8.1}_{-8.1}}$ & $61.1{\scriptstyle^{+17.3}_{-16.4}}$ & $45.9{\scriptstyle^{+18.2}_{-17.7}}$ & $30.2{\scriptstyle^{+10.6}_{-10.2}}$ \\
    \midrule
    \multirow{3}{*}{Qwen3-235B} & Few-Shot & $0.7{\scriptstyle^{+2.2}_{-0.7}}$ & $22.9{\scriptstyle^{+9.6}_{-7.9}}$ & $37.8{\scriptstyle^{+19.1}_{-18.0}}$ & $24.4{\scriptstyle^{+15.7}_{-13.8}}$ & $38.6{\scriptstyle^{+10.0}_{-10.1}}$ \\
     & Zero-Shot & $8.7{\scriptstyle^{+11.3}_{-8.7}}$ & $4.6{\scriptstyle^{+4.2}_{-3.3}}$ & $27.2{\scriptstyle^{+18.6}_{-16.6}}$ & $35.8{\scriptstyle^{+17.7}_{-18.1}}$ & $43.5{\scriptstyle^{+11.0}_{-10.2}}$ \\
     & Direct & $32.1{\scriptstyle^{+16.5}_{-15.0}}$ & $12.8{\scriptstyle^{+7.2}_{-6.6}}$ & $66.5{\scriptstyle^{+18.5}_{-18.9}}$ & $53.1{\scriptstyle^{+18.3}_{-17.9}}$ & $25.7{\scriptstyle^{+9.6}_{-9.1}}$ \\
    \bottomrule
  \end{tabular}
  \caption{%
    Main results across five domains.
    \textbf{SARA Numeric} and \textbf{Airline} report accuracy;
    \textbf{SARA Binary}, \textbf{USCIS}, and \textbf{Housing} report macro~F1.
    Values are reported as mean$^{+\text{upper}}_{-\text{lower}}$, where \textit{upper} and \textit{lower} denote distances from the mean to the 97.5th and 2.5th bootstrap percentiles, respectively.}
  \label{tab:main-results}
\end{table*}

\section{Experimental Setup}
\label{experimental_setup}
We evaluate a diverse set of frontier and open-source language models on \deontic, which requires long-context reasoning over executable rule systems: GPT-4.1, GPT-5.1, GPT-5.2, O3, Claude 4.5 Sonnet, Gemini 2.5 Flash, Kimi K2 Instruct 0905, and Qwen3-235B-A22B-Instruct \citep{singh2025openai, openai2025o3o4mini, team2025kimi, qwen3}. Additional evaluation details are provided in Appendix~\ref{appendix:additional-metrics-bootstrap}.

\subsection{Prompting Strategies}
We evaluate three prompting strategies: \textbf{Direct}, \textbf{Zero-shot}, and \textbf{Few-shot}. Direct prompting predicts the final answer without generating Prolog. Zero-shot prompting generates a complete executable Prolog program from statutes and case facts under domain-specific output constraints for automatic execution and scoring. Few-shot prompting uses the same program-synthesis format with in-context exemplars. Full prompt templates and formatting constraints are provided in Appendix~\ref{appendix:additional-experiment-zero-shot-prompts}.

\subsection{Metrics}
\label{sec:setup-metrics}
For SARA Numeric and Airline, we report accuracy with a \$1 tolerance on the solver output. For SARA Binary, Housing, and USCIS-AAO, we report macro-F1. We also track two error categories: incorrect predictions and \emph{abstentions} (non-executable Prolog or unparsable direct outputs) \citep{jurayj2026language}, and treat abstentions as incorrect during scoring.

We report results on domain-specific hard subsets with 95\% bootstrap confidence intervals. For each case, we sample $K$ generations to capture output stochasticity ($K{=}4$ for SARA Numeric and Binary, and USCIS-AAO; $K{=}3$ for Airline and Housing). Full metric definitions, abstention mapping rules, and the bootstrap procedure are provided in Appendix~\ref{appendix:additional-metrics-bootstrap}.

\section{Results}
\label{sec:results}
We evaluate language models on \deontic hard sets across multiple domains and prompting strategies, focusing on both overall task performance and systematic failure modes. We also analyze model behavior through a detailed error taxonomy.

\begin{figure*}[t]
\centering

\includegraphics[width=0.6\textwidth]{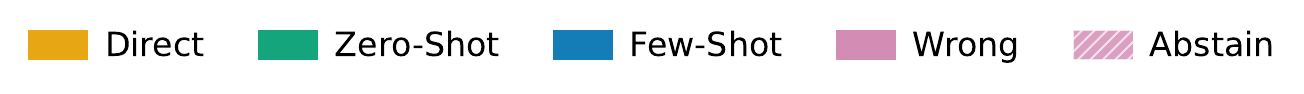}

\vspace{4pt}

\begin{subfigure}[t]{\textwidth}
\centering
\includegraphics[width=\textwidth]{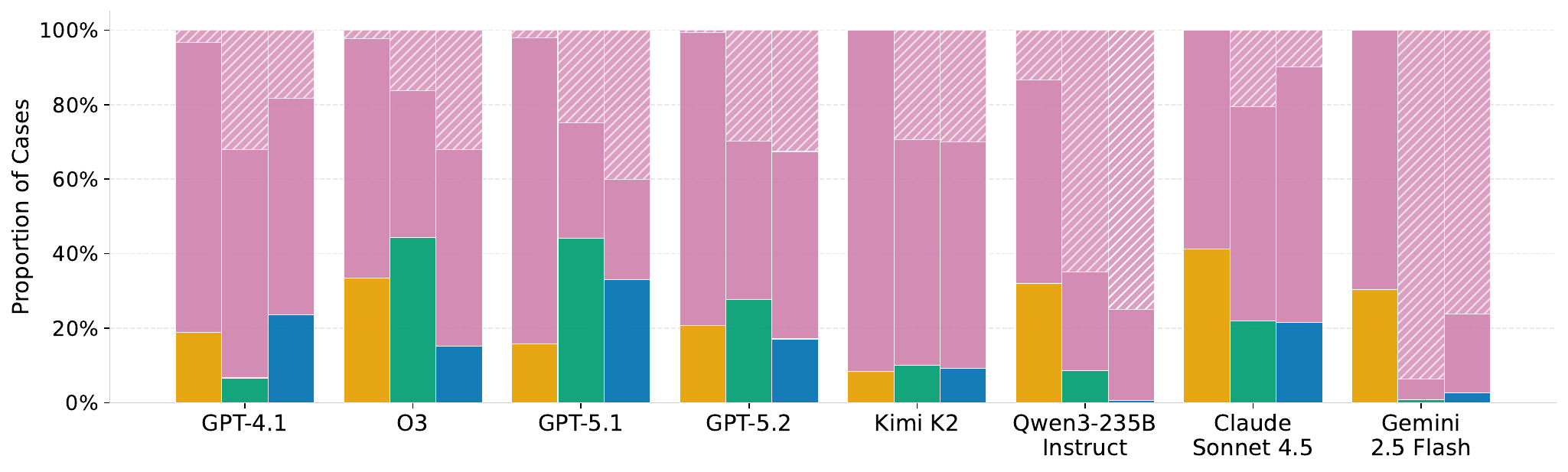}
\caption{SARA Numeric}
\end{subfigure}

\vspace{4pt}

\begin{subfigure}[t]{\textwidth}
\centering
\includegraphics[width=\textwidth]{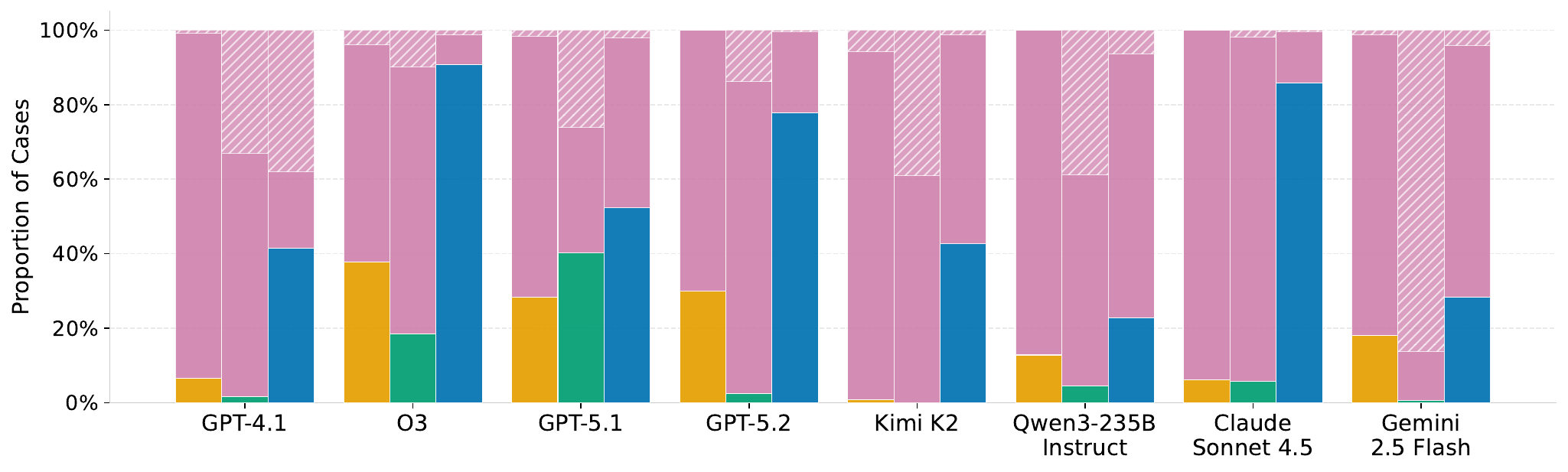}
\caption{Airline}
\end{subfigure}

\caption{%
  Performance decomposition analysis for SARA Numeric and Airline.
  Each model shows three bars (left to right: Direct, Zero-Shot, Few-Shot). Plots for other domains are  in Figure \ref{fig:error-analysis-appendix} (Appendix \ref{appendix:additional-error-bar}).
}
\label{fig:error-analysis-main}
\end{figure*}

\subsection{Main Results}

\paragraph{Reasoning over rules remains challenging for language models.} As Table \ref{tab:main-results} shows, performance varies by domain and prompting strategy: few-shot helps Airline but not SARA Numeric, while zero-shot and direct solving often work better on binary tasks (SARA Binary, USCIS-AAO) than on Housing.

Stronger frontier models (e.g., GPT-5.1, O3, Claude Sonnet 4.5) perform best overall, but none is consistently best across domains or settings. Notably, confidence intervals remain wide across hard subsets, in both smaller hard subsets (e.g., SARA Numeric) and larger ones (e.g., Airline and Housing), suggesting variability beyond sample-size effects.  

\paragraph{There is a clear gap between frontier and open-source models.}
Open-source models perform worse in few-shot and zero-shot settings and are more prompt-sensitive: Qwen3-235B, for instance, rises from near-random few-shot performance (0.7 on SARA Numeric) to 32.1 under direct prompting. The gap narrows on binary tasks, where direct prompting yields competitive scores (66.5 on SARA Binary for Qwen3-235B and 68.4 for Kimi K2), but open-source models remain less reliable for exact rule-based computation.

\paragraph{Performance decomposition reveals systematic failure modes.}
We analyze the proportion of correct predictions, incorrect answers, and abstentions for each model and prompting strategy, as shown in Figure~\ref{fig:error-analysis-main}.

Across all the domains, a large fraction of errors arise from incorrect answers rather than abstentions. In particular, many models produce executable but incorrect Prolog programs, or return wrong answers under direct prompting. However, when solving tasks via Prolog, models exhibit higher abstention rates, whereas direct prompting reduces abstentions but increases the frequency of incorrect predictions. This highlights a trade-off between coverage and reliability. These results motivate confidence-aware inference. With fine-grained abstention calibration, a smaller model can handle high-confidence cases, while uncertain cases are deferred to a stronger model. This improves cost efficiency and reduces risk in high-stakes domains; developing confidence signals remains future work.

\paragraph{Additional analyses.}
Extended results for coding LLMs and reasoning-effort ablations are detailed in the appendix (Appendix~\ref{appendix:coding-llm-results}; Appendix~\ref{appendix:reasoning-effort-main}).

\subsection{Performance of Locally Trained Models}
\label{sec:resutls-locally-trained-models}
\begin{table}[t]
  \centering
  \footnotesize
  \renewcommand{\arraystretch}{1.15}
  \setlength{\tabcolsep}{5pt}
  \begin{tabular}{@{}ll ccccc@{}}
    \toprule
    & & \multicolumn{2}{c}{\textbf{Accuracy}} & \multicolumn{3}{c}{\textbf{Macro F1}} \\
    \cmidrule(lr){3-4} \cmidrule(lr){5-7}
    \textbf{Model} & \textbf{Setting}
      & \textbf{SARA Num.} & \textbf{Airline}
      & \textbf{SARA Bin.} & \textbf{USCIS-AAO} & \textbf{Housing} \\
    \midrule
    \multirow{3}{*}{Qwen2.5-32B (base)} & Few-Shot & $1.3{\scriptstyle^{+4.4}_{-1.3}}$ & $15.9{\scriptstyle^{+7.9}_{-7.1}}$ & $29.1{\scriptstyle^{+17.3}_{-15.2}}$ & $10.3{\scriptstyle^{+10.7}_{-10.3}}$ & $35.1{\scriptstyle^{+10.1}_{-8.8}}$ \\
     & Zero-Shot & $0.7{\scriptstyle^{+5.0}_{-0.7}}$ & $0.0{\scriptstyle^{+0.0}_{-0.0}}$ & $18.2{\scriptstyle^{+13.9}_{-11.5}}$ & $42.8{\scriptstyle^{+17.5}_{-17.9}}$ & $30.9{\scriptstyle^{+10.5}_{-9.7}}$ \\
     & Direct & $4.4{\scriptstyle^{+7.1}_{-4.4}}$ & $0.9{\scriptstyle^{+2.9}_{-0.9}}$ & $62.8{\scriptstyle^{+15.8}_{-17.7}}$ & $36.1{\scriptstyle^{+17.4}_{-18.4}}$ & $24.3{\scriptstyle^{+8.9}_{-8.4}}$ \\
    \midrule
    \multirow{3}{*}{SFT} & Few-Shot & $1.3{\scriptstyle^{+4.4}_{-1.3}}$ & $53.5{\scriptstyle^{+11.5}_{-11.0}}$ & $41.6{\scriptstyle^{+18.7}_{-17.5}}$ & $35.0{\scriptstyle^{+18.3}_{-17.3}}$ & $42.0{\scriptstyle^{+10.5}_{-11.3}}$ \\
     & Zero-Shot & $2.2{\scriptstyle^{+6.4}_{-2.2}}$ & $0.0{\scriptstyle^{+0.0}_{-0.0}}$ & $26.4{\scriptstyle^{+18.2}_{-15.8}}$ & $48.5{\scriptstyle^{+18.3}_{-18.6}}$ & $40.5{\scriptstyle^{+10.7}_{-10.1}}$ \\
     & Direct & $9.9{\scriptstyle^{+10.1}_{-9.9}}$ & $0.9{\scriptstyle^{+2.9}_{-0.9}}$ & $64.7{\scriptstyle^{+17.2}_{-18.4}}$ & $55.8{\scriptstyle^{+18.4}_{-18.5}}$ & $32.3{\scriptstyle^{+9.7}_{-8.2}}$ \\
    \midrule
    \multirow{3}{*}{DPO + SFT} & Few-Shot & $4.1{\scriptstyle^{+7.3}_{-4.1}}$ & $58.0{\scriptstyle^{+10.8}_{-10.5}}$ & $43.1{\scriptstyle^{+17.5}_{-18.2}}$ & $45.7{\scriptstyle^{+17.9}_{-17.5}}$ & $44.2{\scriptstyle^{+10.9}_{-11.3}}$ \\
     & Zero-Shot & $1.4{\scriptstyle^{+4.3}_{-1.4}}$ & $0.3{\scriptstyle^{+2.2}_{-0.3}}$ & $41.6{\scriptstyle^{+17.8}_{-16.7}}$ & $46.7{\scriptstyle^{+17.5}_{-18.4}}$ & $43.3{\scriptstyle^{+10.5}_{-10.4}}$ \\
     & Direct & $7.1{\scriptstyle^{+10.0}_{-7.1}}$ & $0.7{\scriptstyle^{+1.8}_{-0.7}}$ & $61.1{\scriptstyle^{+17.3}_{-18.6}}$ & $54.2{\scriptstyle^{+17.2}_{-19.0}}$ & $28.8{\scriptstyle^{+9.0}_{-8.2}}$ \\
    \midrule
    \multirow{3}{*}{GRPO + SFT} & Few-Shot & $5.4{\scriptstyle^{+8.9}_{-5.4}}$ & $55.2{\scriptstyle^{+11.0}_{-10.2}}$ & $44.0{\scriptstyle^{+19.6}_{-19.1}}$ & $40.8{\scriptstyle^{+17.3}_{-16.7}}$ & $41.0{\scriptstyle^{+11.2}_{-10.4}}$ \\
     & Zero-Shot & $6.2{\scriptstyle^{+8.1}_{-6.2}}$ & $0.3{\scriptstyle^{+1.0}_{-0.3}}$ & $51.6{\scriptstyle^{+18.9}_{-19.5}}$ & $40.3{\scriptstyle^{+19.1}_{-16.2}}$ & $43.9{\scriptstyle^{+11.0}_{-10.6}}$ \\
     & Direct & $8.9{\scriptstyle^{+11.1}_{-8.9}}$ & $0.9{\scriptstyle^{+2.9}_{-0.9}}$ & $56.0{\scriptstyle^{+18.8}_{-18.7}}$ & $55.8{\scriptstyle^{+18.4}_{-18.5}}$ & $31.8{\scriptstyle^{+9.5}_{-8.0}}$ \\
    \midrule
    \multirow{3}{*}{GRPO + DPO + SFT} & Few-Shot & $4.8{\scriptstyle^{+9.5}_{-4.8}}$ & $60.4{\scriptstyle^{+10.8}_{-10.4}}$ & $54.0{\scriptstyle^{+20.2}_{-19.6}}$ & $46.4{\scriptstyle^{+17.9}_{-18.2}}$ & $43.0{\scriptstyle^{+10.7}_{-10.7}}$ \\
     & Zero-Shot & $3.6{\scriptstyle^{+7.8}_{-3.6}}$ & $0.0{\scriptstyle^{+0.0}_{-0.0}}$ & $46.4{\scriptstyle^{+17.7}_{-18.2}}$ & $48.4{\scriptstyle^{+18.8}_{-18.4}}$ & $42.3{\scriptstyle^{+11.4}_{-10.8}}$ \\
     & Direct & $6.5{\scriptstyle^{+10.7}_{-6.5}}$ & $0.3{\scriptstyle^{+2.2}_{-0.3}}$ & $59.2{\scriptstyle^{+18.9}_{-19.1}}$ & $54.2{\scriptstyle^{+17.2}_{-19.0}}$ & $28.5{\scriptstyle^{+8.8}_{-8.1}}$ \\
    \bottomrule
  \end{tabular}
  \caption{%
    Results for locally trained models across five domains.
    \textbf{SARA Numeric} and \textbf{Airline} report accuracy;
    \textbf{SARA Binary}, \textbf{USCIS-AAO}, and \textbf{Housing} report macro~F1.
    Values are reported as mean$^{+\text{upper}}_{-\text{lower}}$, where \textit{upper} and \textit{lower} denote distances from the mean to the 97.5th and 2.5th bootstrap percentiles, respectively.}
  \label{tab:trained-model-results}
\end{table}

Table \ref{tab:trained-model-results} reports performance on the hard subsets of \deontic when fine-tuning Qwen2.5-32B-Instruct \citep{qwen2.5}. We ensure that hard-set examples are excluded from training, as described in \S\ref{sec:hard-set-construction}. Following Unilaw-R1 \citep{cai2025unilaw} and OLMO 3 training recipe \citep{olmo2025olmo}, we apply Supervised Fine-Tuning (SFT), Direct Preference Optimization (DPO) \citep{rafailov2023direct}, and Dr.~GRPO \citep{liu2025understanding}, a variant of Group Relative Policy Optimization (GRPO) \citep{shao2024deepseekmath} with a predicate-aware reward function inspired by \citet{chen2025predicate}. Additional training details are provided in Appendix~\ref{appendix:additional-locally-trained-models}.

\paragraph{Fine-tuning improves classification but not numeric reasoning.}
SFT improves performance over the base model, especially on binary tasks such as SARA Binary and USCIS-AAO (e.g., few-shot USCIS-AAO rises from 10.3 to over 45 with alignment-based training). By contrast, SARA Numeric remains below 10 across methods and prompting strategies, indicating that precise rule-based computation is still a key bottleneck.

\paragraph{Reinforcement Learning provides additional gains, but inconsistently.}
DPO and Dr. GRPO generally outperform SFT alone, especially in few-shot settings. For instance, GRPO+DPO+SFT achieves the strongest performance on Airline and SARA Binary. However, these gains are not uniform across domains or prompting strategies, and improvements are limited on more challenging reasoning tasks.

\paragraph{Prompt sensitivity persists after training.}
After post-hoc training, models remain highly sensitive to prompting strategy. Few-shot prompting consistently yields the best performance on Airline, while direct prompting is more effective for binary reasoning tasks. Overall, while SFT and RL improve performance, current RLHF-based methods are insufficient for robust rule-grounded reasoning on \deontic. Developing methods that reliably produce correct, executable Prolog remains an important direction for future work.

\begin{table}[t]
\centering
\small
\setlength{\tabcolsep}{6pt}
\begin{tabular}{llcccc}
\toprule
\textbf{Domain} & \textbf{Prompt} & \textbf{Wrong Rule} & \textbf{Entity / Fact} & \textbf{Numerical} & \textbf{Impl.\ Bugs} \\
\midrule
\multirow{2}{*}{SARA Numeric}
  & Few-Shot  & 31\% & \textbf{42\%} & 22\% & 6\% \\
  & Zero-Shot & 35\% & \textbf{52\%} & 13\% & --- \\
\midrule
\multirow{2}{*}{SARA Binary}
  & Few-Shot  & \textbf{55\%} & 45\% & --- & --- \\
  & Zero-Shot & \textbf{47\%} & 42\% & 12\% & --- \\
\midrule
\multirow{2}{*}{Airline}
  & Few-Shot  & --- & --- & \textbf{100\%} & --- \\
  & Zero-Shot & 12\% & 6\% & \textbf{75\%} & 7\% \\
\midrule
\multirow{2}{*}{Housing}
  & Few-Shot  & \textbf{96.8\%} & --- & --- & 3.2\% \\
  & Zero-Shot & \textbf{93.5\%} & --- & --- & 6.5\% \\
\midrule
\multirow{2}{*}{USCIS-AAO}
  & Few-Shot  & \textbf{42.6\%} & 11.5\% & --- & 42.3\% \\
  & Zero-Shot & \textbf{77.3\%} & --- & --- & 22.7\% \\
\bottomrule
\end{tabular}
\caption{
Distribution of error types across domains and prompting strategies,
based on hard cases where GPT-5.1, GPT-5.2, and O3 failed on every $K$ generation.
Percentages are normalized per row; the dominant category per row is \textbf{bolded}.
\textbf{Wrong Rule}: the model applied the wrong statute, subsection, or fee schedule.
\textbf{Entity / Fact}: correct rule but wrong facts encoded (e.g., filing status, dependency relationships, bag type).
\textbf{Numerical}: correct rule and facts but arithmetic or threshold errors (e.g., phaseout computation, bag count optimization).
\textbf{Impl.\ Bugs}: low-level Prolog programming errors unrelated to reasoning.
}
\label{tab:error_types}
\end{table}

\subsection{Failure Modes in Prolog Reasoning} 

By comparing model-generated Prolog and ground truth Prolog, we categorize errors on the hard cases made by frontier models (GPT-5.1, GPT-5.2, O3) into four types: \emph{Wrong Rule}, \emph{Entity / Fact}, \emph{Numerical}, and \emph{Implementation Bugs}. Table~\ref{tab:error_types} reports the distribution of these error types across domains and prompting strategies. 

We can observe that \textbf{rule selection errors} dominate in legally complex settings such as Housing and USCIS-AAO, where models frequently apply incorrect statutes or misinterpret eligibility criteria. This suggests that retrieving and grounding the correct legal rule remains a primary bottleneck, even when models can generate syntactically valid Prolog. Second, \textbf{entity and fact extraction errors} are particularly prominent in SARA tasks, especially under zero-shot prompting. In these cases, models often identify the correct rule but incorrectly encode case-specific details (e.g., filing status or dependency relationships), leading to incorrect outcomes despite otherwise valid reasoning structure. 

Overall, these results highlight that failures are \emph{domain-dependent}: legal domains are bottlenecked by rule selection, structured reasoning tasks by fact extraction, and quantitative domains by arithmetic precision. Improving Prolog-based reasoning systems will therefore require targeted advances in rule grounding, structured information extraction, and reliable numerical computation, rather than a single unified fix.

\section{Related Work}
\subsection{Related Datasets for Deontic Reasoning}

Rule-reasoning benchmarks have expanded across formal and natural-language settings: FOLIO \citep{han2024folio}, LogicBench \citep{parmar2024logicbench}, and JustLogic \citep{chen2025justlogic} emphasize formal structure and representation alignment, while ProofWriter \citep{tafjord2021proofwriter} and ProntoQA \citep{saparov2022language} focus on multi-step entailment and synthetic first-order-logic reasoning with verifiable proof chains. Recent process-oriented benchmarks include LOGicalThought \citep{nananukul2025logicalthought}, which targets negation, implication, and defeasible reasoning, and ContextBench \citep{li2026contextbench}, which evaluates intermediate context retrieval for coding agents using human-annotated gold contexts.

RuleArena \citep{zhou2025rulearena} and CL-bench \citep{dou2026clbenchbenchmarkcontextlearning} are the closest to our setting. Compared with RuleArena and CL-bench, \deontic is larger (6,232 vs.\ 816 tasks in RuleArena), covers additional high-stakes legal domains, and provides executable Prolog artifacts for program-trace analysis. CL-bench refers to this setup as \textit{context learning}; related formulations also appear in prior statutory-reasoning work \citep{holzenberger-van-durme-2023-connecting, blair2023openai}. Like CL-bench, \deontic minimizes reliance on parametric knowledge by providing all required statutes and facts in each instance.

\subsection{Symbolic Reasoning in Large Language Models}

Recent work on LLM reasoning integrates structured symbolic methods, often referred to as autoformalization \citep{wu2022autoformalization, mensfelt2026towards}. A common neuro-symbolic line combines neural modules with probabilistic logic for differentiable or approximate inference \citep{van2023nesi, pryor2022neupsl, huang2021scallop}. Another common approach uses a two-stage pipeline: translate natural language into symbolic programs, then execute them with external solvers. Faithful Chain-of-Thought \citep{lyu2023faithful} and Logic-LM \citep{pan2023logic} show that executable formalisms such as PDDL and Prolog can improve faithfulness. LOGicalThought \citep{nananukul2025logicalthought} similarly maps natural language into symbolic representations and reasons with an external solver (ErgoAI). 

More recent work studies direct Prolog generation. For example, prior studies show that LLMs can generate Prolog programs for arithmetic reasoning \citep{yang2405arithmetic}. In legal reasoning, \citet{jurayj2026language} show that combining LLMs with Prolog solvers improves statutory reasoning. Our work builds on these directions and provides Prolog reference for all the instances.

\section{Conclusion}
We introduce \deontic, a benchmark of 6,232 high-stakes, long-context rule-reasoning tasks spanning U.S. federal taxes, airline baggage policies, U.S. immigration administration, and U.S. state housing law. \deontic supports both direct language reasoning and solver-assisted execution with reference Prolog programs. Across frontier reasoning and coding models, performance remains weak on difficult subsets, and added reasoning effort does not consistently help. Although SFT and RL improve performance, reliable end-to-end gains remain limited.

\section*{Ethics Statement}
This benchmark studies model behavior in high-stakes legal and policy domains, where errors can cause harm if systems are deployed without appropriate human oversight. We release \deontic strictly for research and evaluation purposes, not for operational decision-making. Model outputs should not be treated as professional legal, tax, or policy advice.

\section*{Acknowledgements}
This work was funded in part by the Defense Advanced Research Projects Agency (DARPA) CODORD program, and by Schmidt Sciences. Any opinions, findings, and conclusions or recommendations expressed in this material are those of the author(s) and do not necessarily reflect the views of DARPA or Schmidt Sciences.

\bibliography{colm2026_conference}
\bibliographystyle{colm2026_conference}

\newpage
\appendix
\section{Additional Prompts}
\label{appendix:additional-prompts}
\paragraph{Additional Instructions for Facts Extraction.}
The following instruction is used to prompt GPT-5-mini to extract factual information from the content of USCIS cases:

\begin{promptbox}{USCIS facts extraction prompt}
\begin{Verbatim}[fontsize=\small,breaklines=true,breakanywhere=true]
Use this administrative appeals office appeal case to extract only the facts related to the petitioner (not the analysis of the administrative appeals office director) and the reason of appealing. Avoid including any application of the law. Make it a description of only the facts and the reason of appealing. It should not contain any analysis of the described rules or other rules or precedents that are not strictly facts stated by the petitioner. This description of facts should be written in natural language and coherent when reading. It shouldn't be only a list of facts but a description of the facts as if it where a human description of them. Avoid stating the conclusion of the case or what was resolved. The opinion and resolution of the administrative appeals office must not be included. Case text under analysis: {case_text}
\end{Verbatim}
\end{promptbox}

\section{Additional Experiment Setup}
\label{appendix:additional-experiment-setup}

\subsection{Prolog Generation Pipeline Details}
\label{appendix:prolog-generation-details}

We consider two LLM-to-Prolog generation pipelines. In the first, we pre-translate each unique statute into a Prolog module, compile-check it under SWI-Prolog \citep{wielemaker:2011:tplp}, and store it as a reusable library. For each question, the model then generates only the question program that calls precompiled statute predicates. In the second, we use OpenAI o3 \citep{openai2025o3o4mini} (medium reasoning effort) to generate statute modules on the fly, compile-check them, and then generate the question program against those interfaces. This avoids maintaining a global library and better supports frequently changing, jurisdiction-dependent regulations.

In our experiments, we use the second pipeline. This design localizes failures to a single statute/question artifact pair and allows successful generations to transfer cleanly to updated corpora without maintaining a monolithic global library.

Each generated artifact (statute modules + question program) is executed and counted as passing only if (1) it compiles and runs cleanly in SWI-Prolog and (2) the emitted answer matches the gold label. Failures include incorrect answers, missing or extra output, missing code, module import errors, Prolog warnings, and execution timeouts (20s). During curation, we retain only $<$Statutes + Case Facts, Prolog$>$ pairs with passing artifacts.

We first generate an initial batch of Prolog programs for a small subset in each domain, then send a random subset to human experts for validation. Curation proceeds with a checklist: annotators verify (1) rule coverage (all legally relevant conditions are encoded), (2) fact fidelity (case facts are preserved without label leakage), (3) executability of Prolog programs (no warnings/import/runtime issues), and (4) answer agreement with the gold label; artifacts that fail are regenerated, and each edited artifact is re-executed before acceptance. After curation, we stratify accepted examples by domain, sample a small balanced set as few-shot demonstrations (mixing positive/negative outcomes and common edge cases), and regenerate the remaining instances using this curated few-shot examples.


To increase the number of passing artifacts, we allow up to two attempts per instance: on failure, we re-prompt with the previous code and SWI-Prolog compiler/runtime output; if the second attempt fails, the instance is discarded. This mirrors tool-integrated evaluation settings where generated programs are executed and verified by external systems, yielding deterministic feedback signals rather than purely textual judgments \citep{yang2023leandojotheoremprovingretrievalaugmented}.

For USCIS-AAO, we apply additional syntax-focused prompt constraints to reduce parsing and compilation errors, and we encourage a static final proof clause when possible. Full prompts and query templates are provided in Appendix~\ref{appendix:additional-prompts} and Appendix~\ref{appendix:additional-experiment-zero-shot-prompts}.

\subsection{Detailed Hard Set Construction}
\label{appendix:additional-hard-set-construction}

We identify candidate hard instances by running three frontier models---OpenAI o3, GPT-5.2, and Claude 4.5 Sonnet---on every \deontic problem with Prolog generation, using two runs per model. Any instance for which at least one generation fails is marked as \emph{potentially hard}. We then conduct iterative human review to remove noisy or ambiguous cases and retain only genuinely challenging instances.

After this process, we do not place all validated hard instances into the final hard subset. Instead, we partition these validated hard problems into two portions: one is kept as the final \emph{hard set} for evaluation, and the remainder is returned to the full dataset for training and related experiments. This strategy keeps evaluation focused while ensuring that training still includes challenging examples rather than being dominated by easier cases.

\subsection{Detailed Metrics and Bootstrap Procedure}
\label{appendix:additional-metrics-bootstrap}

\paragraph{Accuracy.}
For SARA Numeric and Airline, a prediction is correct if the value obtained from SWI-Prolog is within \$1 of the ground truth. Formally, given model output $y$ and ground truth $\hat{y}$, correctness is defined as $|\mathrm{round}(y) - \mathrm{int}(\hat{y})| \leq 1$.

\paragraph{Macro-F1.}
For SARA Binary, Housing, and USCIS-AAO, we report macro-F1. Given confusion matrix counts TP, FP, FN, and TN, the class-wise scores are $F_1^{(1)} = \frac{2\mathrm{TP}}{2\mathrm{TP} + \mathrm{FP} + \mathrm{FN}}$ and $F_1^{(0)} = \frac{2\mathrm{TN}}{2\mathrm{TN} + \mathrm{FN} + \mathrm{FP}}$, and macro-F1 is $\frac{F_1^{(1)} + F_1^{(0)}}{2}$.

\paragraph{Abstentions and scoring.}
We distinguish two failure types: (i) incorrect predictions from usable outputs, and (ii) \emph{abstentions}, where the model output cannot be used for scoring (e.g., non-executable Prolog or unparsable direct output). Abstentions are counted as incorrect predictions. For SARA Numeric and Airline, abstentions are assigned \$-1. For SARA Binary, Housing, and USCIS-AAO, abstentions are mapped to the opposite class before computing the confusion matrix.

\paragraph{Bootstrap confidence intervals.}
All main-table metrics are computed on hard subsets. To estimate uncertainty, we use bootstrap resampling with $B=1000$ replicates. In each replicate, we sample $N$ cases with replacement (where $N$ is the hard-subset size), then sample one of the $K$ outputs per case uniformly at random to account for output stochasticity. We report 95\% confidence intervals using the empirical 2.5th and 97.5th percentiles. We use $K=4$ for SARA Numeric, SARA Binary, and USCIS-AAO, and $K=3$ for Airline and Housing.

\subsection{Zero-Shot Prompt Templates}
\label{appendix:additional-experiment-zero-shot-prompts}

This appendix presents the full prompt templates used in the \textbf{Zero-Shot} (Standalone Prolog)
condition for each domain.
Placeholders \ph{like this} are replaced at inference time with the corresponding
case-specific content.
All prompts are preceded by the system message:
\textit{``You are a helpful assistant trained to conduct deontic reasoning.''}

\subsection*{SARA Numeric}

\begin{promptbox}{SARA Numeric --- Zero-Shot Prompt Template}
\begin{ttfamily}\small
\#\# Statutes:\\
\ph{statutes}\\[4pt]
\#\# Case: \ph{case text}\\[4pt]
\#\# Question: \ph{question}\\[6pt]
The question above asks about the case in relation to the statutes.
First, write a logic program in prolog to encode the relevant rules defined in the statutes.
Then write a logic program in prolog to encode the facts and rules contained in the case above.
Then write a query to compute and print the value the question asks.\\
You must indicate your prolog code using:\\[2pt]
\hspace*{1em}\textasciigrave\textasciigrave\textasciigrave prolog\\
\hspace*{1em}<YOUR\_LOGIC\_PROGRAM\_HERE>\\
\hspace*{1em}\textasciigrave\textasciigrave\textasciigrave\\[4pt]
For instance, to answer ``How much tax does Samuel owe in 1992'', your final lines should be\\[2pt]
\hspace*{1em}\textasciigrave\textasciigrave\textasciigrave prolog\\
\hspace*{1em}:- tax("Samuel", 1992, Tax), format('Tax result: \textasciitilde{}w\textasciitilde{}n', [Tax]).\\
\hspace*{1em}:- halt.\\
\hspace*{1em}\textasciigrave\textasciigrave\textasciigrave
\end{ttfamily}
\end{promptbox}

\subsection*{SARA Binary}

\begin{promptbox}{SARA Binary --- Zero-Shot Prompt Template}
\begin{ttfamily}\small
\#\# Statutes:\\
\ph{statutes}\\[4pt]
\#\# Case: \ph{case text}\\[4pt]
\#\# Claim: \ph{claim}\\[6pt]
The claim above makes a statement about the case in relation to the statutes.
Determine whether the claim is correct (Entailment) or incorrect (Contradiction).\\[4pt]
First, write a logic program in Prolog to encode the relevant rules defined in the statutes.
Then write a logic program in Prolog to encode the facts contained in the case above.
Then write a query that checks whether the claim holds, and prints the result.\\
You must indicate your prolog code using:\\[2pt]
\hspace*{1em}\textasciigrave\textasciigrave\textasciigrave prolog\\
\hspace*{1em}<YOUR\_LOGIC\_PROGRAM\_HERE>\\
\hspace*{1em}\textasciigrave\textasciigrave\textasciigrave\\[4pt]
Your program's final output must be exactly one of:\\
\hspace*{2em}Result: Entailment\\
\hspace*{2em}Result: Contradiction\\[4pt]
For example, your final lines should follow this pattern:\\[2pt]
\hspace*{1em}\textasciigrave\textasciigrave\textasciigrave prolog\\
\hspace*{1em}:- ( \phantom{x} <your\_verification\_goal>\\
\hspace*{2em}-> format('Result: Entailment\textasciitilde{}n')\\
\hspace*{2em};  format('Result: Contradiction\textasciitilde{}n')\\
\hspace*{2em}).\\
\hspace*{1em}:- halt.\\
\hspace*{1em}\textasciigrave\textasciigrave\textasciigrave\\[4pt]
Follow this format exactly. Think step-by-step before answering.
\end{ttfamily}
\end{promptbox}

\subsection*{Airline}

\begin{promptbox}{Airline --- Zero-Shot Prompt Template}
\begin{ttfamily}\small
\#\# Statutes:\\
\ph{statutes}\\[4pt]
\#\# Case: \ph{case text}\\[4pt]
\#\# Question: \ph{question}\\[6pt]
Write a complete runnable SWI-Prolog program that computes the answer.
Indicate your prolog code using:\\[2pt]
\hspace*{1em}\textasciigrave\textasciigrave\textasciigrave prolog\\
\hspace*{1em}<YOUR\_LOGIC\_PROGRAM\_HERE>\\
\hspace*{1em}\textasciigrave\textasciigrave\textasciigrave\\[4pt]
Final lines should print total result and halt.
\end{ttfamily}
\end{promptbox}

\subsection*{Housing}

\begin{promptbox}{Legal IR --- Zero-Shot Prompt Template}
\begin{ttfamily}\small
You are an expert US housing lawyer and senior SWI-Prolog engineer.
Write a self-contained SWI-Prolog program that translates statutes and question facts.\\[6pt]
\#\#\# HousingQA Sample\\
- state: \ph{state}\\
- focus\_year: 2021\\[4pt]
\#\#\# Question\\
\ph{question}\\[4pt]
\#\#\# Natural Language Statutes\\
\ph{statutes}\\[6pt]
\#\#\# Output requirements\\
- Output one runnable \textasciigrave\textasciigrave\textasciigrave prolog block only.\\
- Define \texttt{housing\_answer(Result)} where \texttt{Result} is \texttt{yes} or \texttt{no}.\\
- Derive \texttt{Result} via statute predicates; do not hardcode.\\
- The file must run correctly under \texttt{swipl -f file.pl}.\\
- Define \texttt{main/0} and end with exactly \texttt{:- initialization(main, main).}\\
- \texttt{main} must print exactly: \texttt{housing\_answer(\textasciitilde{}w).}
\end{ttfamily}
\end{promptbox}

\subsection*{USCIS-AAO}

\begin{promptbox}{USCIS-AAO --- Zero-Shot Prompt Template}
\begin{ttfamily}\small
You are an expert in the analysis of immigration appeals and you will parse these
Administrative Appeals Office cases into Prolog code.
Parse the facts and rules into Prolog code, ensuring that the program's output is binary:
either \textbf{Dismissed} or \textbf{Accepted}.\\[6pt]
You are given the following facts:\\
\ph{case facts}\\[4pt]
You are given the following rules:\\
\ph{statutory rules}\\[6pt]
For true/false or yes/no predicates, use arity 0 and check it is consistent for all clauses.
For example, use \texttt{there\_is\_evidence.} instead of \texttt{there\_is\_evidence(True)}.\\
Include this structure at the end and work assuming that \texttt{eligibility\_met} is all the conditions necessary to Accept the case. Therefore, this must be the last part of the program.\\
Do not include the query in the prolog output, only include the entrypoint.\\
When executed, print exactly one token: \texttt{Accepted} or \texttt{Dismissed}.\\[4pt]
Include this structure at the end (where \texttt{eligibility\_met} encodes all conditions
necessary to accept the case):\\[4pt]
\hspace*{1em}decision(Result) :-\\
\hspace*{2em}( \phantom{x} eligibility\_met\\
\hspace*{2em}->  Result = 'Accepted'\\
\hspace*{2em};   Result = 'Dismissed'\\
\hspace*{2em}).\\[4pt]
\hspace*{1em}main :-\\
\hspace*{2em}catch(\\
\hspace*{3em}( \phantom{x} decision(Result),\\
\hspace*{4em}writeln(Result)\\
\hspace*{3em}),\\
\hspace*{3em}error(existence\_error(procedure, PI), \_),\\
\hspace*{3em}handle\_undefined(PI)\\
\hspace*{2em}).\\[4pt]
\hspace*{1em}handle\_undefined(Name/Arity) :-\\
\hspace*{2em}( \phantom{x} current\_predicate(Name/OtherArity),\\
\hspace*{3em}OtherArity \textbackslash= Arity\\
\hspace*{2em}->  format('Programming error: called \textasciitilde{}w/\textasciitilde{}w, but only \textasciitilde{}w/\textasciitilde{}w is defined.\textasciitilde{}n',\\
\hspace*{5em}[Name, Arity, Name, OtherArity])\\
\hspace*{2em};   format('Lack of information: predicate \textasciitilde{}w/\textasciitilde{}w is not defined.\textasciitilde{}n',\\
\hspace*{5em}[Name, Arity])\\
\hspace*{2em}).
\end{ttfamily}
\end{promptbox}

\begin{table}[h]
\centering
\begin{tabular}{lcc}
\toprule
\textbf{Parameter} & \textbf{SFT} & \textbf{DPO} \\
\midrule
Base model & Qwen2.5-32B-Instruct & SFT checkpoint \\
Method & LoRA & LoRA \\
LoRA rank & 8 & 8 \\
LoRA target modules & All & All \\
Sequence length & 8{,}000 & 8{,}000 \\
Per-device batch size & 1 & 1 \\
Gradient accumulation steps & 8 & 8 \\
Learning rate & $1 \times 10^{-4}$ & $7 \times 10^{-8}$ \\
LR scheduler & Cosine & Linear \\
Warmup ratio & 0.1 & 0.1 \\
Epochs & 3 & 1 \\
DPO loss & --- & Sigmoid \\
DPO $\beta$ & --- & 2.5 \\
Precision & BF16 & BF16 \\
DeepSpeed stage & ZeRO-2 & ZeRO-3 \\
\bottomrule
\end{tabular}
\caption{Training hyperparameters for SFT and DPO fine-tuning of Qwen2.5-32B-Instruct.}
\label{tab:training-hyperparams}
\end{table}

\subsection{Additional Details of Locally Trained Models}
\label{appendix:additional-locally-trained-models}

In this section we go through more in-depth details of the training pipeline mentioned in section \ref{sec:resutls-locally-trained-models}. Specifically, we train Qwen2.5-32B-Instruct through Supervised Fine-tuning (SFT), Direct Preference Optimization (DPO), and Reinforcement Learning.  

\subsubsection{SFT and DPO}
\paragraph{Data Preparation} In SFT, we formatted the data in Alpaca Format with <Statute + Contexts, Prolog> pairs. For DPO, we need to generate preference dataset, which includes chosen (the ground truth Prolog in \deontic) and, following the delta learning hypothesis \citep{geng2025delta}, we generate rejected answers by using the off-the-shelf base model. 

\paragraph{Supervised Fine-Tuning (SFT)}
In SFT, the model is trained to maximize the likelihood of the ground-truth Prolog program $y$ given the input $x$ (statutes and case context). The objective is the standard negative log-likelihood:
\begin{equation}
\mathcal{L}_{\text{SFT}}(\theta)
= - \mathbb{E}_{(x, y) \sim \mathcal{D}}
\left[
\sum_{t=1}^{T} \log p_{\theta}(y_t \mid x, y_{<t})
\right],
\end{equation}
where $y_t$ denotes the $t$-th token in the target sequence.

\paragraph{Direct Preference Optimization (DPO)}
DPO optimizes the model using preference triples $(x, y^{+}, y^{-})$, where
$y^{+}$ is the preferred output and $y^{-}$ is the dispreferred output.
Following \citet{rafailov2023direct}, the objective is
\begin{equation}
\mathcal{L}_{\mathrm{DPO}}(\theta)
=
-\mathbb{E}_{(x,y^{+},y^{-})\sim\mathcal{D}}
\left[
\log \sigma\left(
\beta \log \frac{\pi_{\theta}(y^{+}\mid x)}{\pi_{\mathrm{ref}}(y^{+}\mid x)}
-
\beta \log \frac{\pi_{\theta}(y^{-}\mid x)}{\pi_{\mathrm{ref}}(y^{-}\mid x)}
\right)
\right],
\end{equation}
where $\sigma(\cdot)$ is the sigmoid function, $\beta$ controls the strength
of the preference optimization, and $\pi_{\mathrm{ref}}$ is the reference
policy.

\paragraph{SFT and DPO Training Configuration}

We use LLaMA Factory \citep{zheng2024llamafactory} to run SFT and DPO with configuration specified in Table \ref{tab:training-hyperparams}. We train these models on 4 H100 GPUs. 

\subsubsection{Reinforcement Learning}
\paragraph{DR.~GRPO}
For reinforcement learning, we adopt Dr.~GRPO \citep{liu2025understanding}, an unbiased variant of GRPO. 
Given an input question $q$, we sample a group of $G$ responses $\{o_1,\dots,o_G\}$ from the old policy $\pi_{\theta_{\mathrm{old}}}(\cdot \mid q)$. 
The clipped surrogate objective is
\begin{equation}
\mathcal{J}_{\mathrm{Dr.\,GRPO}}(\theta)
=
\mathbb{E}_{q \sim p_Q,\{o_i\}_{i=1}^{G}\sim \pi_{\theta_{\mathrm{old}}}(\cdot \mid q)}
\left[
\frac{1}{G}
\sum_{i=1}^{G}
\sum_{t=1}^{|o_i|}
\min\!\left(
r_{i,t}(\theta)\,\hat{A}_{i,t},
\,
\mathrm{clip}\!\left(r_{i,t}(\theta), 1-\epsilon, 1+\epsilon\right)\hat{A}_{i,t}
\right)
\right],
\end{equation}
where
\begin{equation}
r_{i,t}(\theta)
=
\frac{\pi_{\theta}(o_{i,t}\mid q, o_{i,<t})}
{\pi_{\theta_{\mathrm{old}}}(o_{i,t}\mid q, o_{i,<t})},
\end{equation}
and the group-relative advantage is
\begin{equation}
\hat{A}_{i,t}
=
R(q,o_i)
-
\frac{1}{G}\sum_{j=1}^{G} R(q,o_j).
\end{equation}

\paragraph{Reward Function} The scalar reward $R(q, o_i)$ is computed by executing the generated Prolog program with SWI-Prolog (20-second timeout) and comparing its output against a ground-truth label. We define a three-level scheme: 
\paragraph{(1) Correct execution.} The response is parsed for a fenced \texttt{prolog} code block, which is then run through SWI-Prolog. If the printed output matches the ground-truth label exactly, the model receives the full reward: \begin{equation} R(q, o_i) = 1. \end{equation}

\paragraph{(2) Predicate-overlap partial credit.} If the extracted code is \emph{non-executable} (syntax or runtime error), we compare the set of predicate signatures (name/arity pairs, after stripping comments) between the generated program $P$ and the reference program $P^*$. The partial reward is proportional to their Jaccard similarity: \begin{equation} R(q, o_i) = \frac{|S(P) \cap S(P^*)|}{|S(P) \cup S(P^*)|} \cdot \delta, \quad \delta = 0.2, \end{equation} where $S(\cdot)$ denotes the set of predicate signatures extracted from a program. This gives a small signal ($R \in [0, 0.2]$) that rewards structural alignment with the reference even when the code fails to run. 

\paragraph{(3) All other cases.} A reward of $0$ is assigned when no Prolog block is found, the code times out, or the code runs but produces the wrong answer.

\noindent The binary nature of the correctness signal (1 vs.\ 0 for executable outputs) encourages the model to produce Prologs that compute the exact right answer, while the predicate-overlap term provides a weak gradient for non-executable generations to adopt the correct predicates.

\paragraph{Dr.~GRPO Training Config}

Table \ref{tab:grpo-hyperparams} contains the training details of Dr.~GRPO. We use VeRL \citep{sheng2024hybridflow} with 4 H200 GPUs. 

\begin{table}[h]
\centering
\begin{tabular}{lc}
\toprule
\textbf{Parameter} & \textbf{DR.~GRPO} \\
\midrule
Base model & DPO checkpoint \\
Method & LoRA \\
LoRA rank ($r$) & 128 \\
LoRA $\alpha$ & 128 \\
Learning rate & $2 \times 10^{-6}$ \\
Epochs & 5 \\
Train batch size & 128 \\
Max prompt length & 7{,}800 \\
Max response length & 1{,}024 \\
Rollout samples ($G$) & 5 \\
KL penalty & None \\
Entropy coefficient & 0 \\
Advantage normalisation & None \\
Loss aggregation & \texttt{seq-mean-token-sum-norm} \\
\bottomrule
\end{tabular}
\caption{Training hyperparameters for DR.~GRPO training.}
\label{tab:grpo-hyperparams}
\end{table}

\section{Additional Experiment Results} 

\subsection{Performance of Coding LLMs}
\label{appendix:coding-llm-results}
We also evaluate the performance of GPT-5.2-Codex \citep{openai2025gpt52codex}, Qwen3-Coder-Next-FP8, and Qwen/Qwen3-Coder-480B-A35B-Instruct-FP8 \citep{Qwen3-Coder-Next} on \deontic. The results are shown in Table \ref{tab:coding-agent-results}.

\begin{table}[t]
  \centering
  \footnotesize
  \renewcommand{\arraystretch}{1.15}
  \setlength{\tabcolsep}{5pt}
  \begin{tabular}{@{}ll ccccc@{}}
    \toprule
    & & \multicolumn{2}{c}{\textbf{Accuracy}} & \multicolumn{3}{c}{\textbf{Macro F1}} \\
    \cmidrule(lr){3-4} \cmidrule(lr){5-7}
    \textbf{Model} & \textbf{Setting}
      & \textbf{SARA Num.} & \textbf{Airline}
      & \textbf{SARA Bin.} & \textbf{USCIS-AAO} & \textbf{Housing} \\
    \midrule
    \multirow{3}{*}{GPT-5.2-Codex} & Few-Shot & $26.6{\scriptstyle^{+13.4}_{-12.3}}$ & $95.5{\scriptstyle^{+3.2}_{-5.5}}$ & $28.8{\scriptstyle^{+17.0}_{-16.3}}$ & $34.9{\scriptstyle^{+17.1}_{-15.2}}$ & $41.3{\scriptstyle^{+9.9}_{-9.1}}$ \\
     & Zero-Shot & $45.8{\scriptstyle^{+17.1}_{-14.3}}$ & $25.6{\scriptstyle^{+10.6}_{-9.4}}$ & $16.6{\scriptstyle^{+13.4}_{-13.2}}$ & $47.4{\scriptstyle^{+18.3}_{-17.4}}$ & $31.3{\scriptstyle^{+10.1}_{-9.8}}$ \\
     & Direct & $29.5{\scriptstyle^{+13.3}_{-15.2}}$ & $47.0{\scriptstyle^{+10.5}_{-10.7}}$ & $36.5{\scriptstyle^{+18.5}_{-16.8}}$ & $52.7{\scriptstyle^{+19.4}_{-18.2}}$ & $17.5{\scriptstyle^{+8.5}_{-7.8}}$ \\
    \midrule
    \multirow{3}{*}{Qwen3-Coder-Next} & Few-Shot & $0.0{\scriptstyle^{+0.0}_{-0.0}}$ & $0.0{\scriptstyle^{+0.0}_{-0.0}}$ & $28.9{\scriptstyle^{+16.9}_{-15.1}}$ & $16.4{\scriptstyle^{+13.5}_{-13.0}}$ & $36.9{\scriptstyle^{+11.3}_{-10.0}}$ \\
     & Zero-Shot & $0.0{\scriptstyle^{+0.0}_{-0.0}}$ & $0.0{\scriptstyle^{+0.0}_{-0.0}}$ & $21.7{\scriptstyle^{+15.6}_{-15.1}}$ & $24.5{\scriptstyle^{+14.8}_{-12.0}}$ & $32.8{\scriptstyle^{+10.8}_{-10.5}}$ \\
     & Direct & $4.3{\scriptstyle^{+7.1}_{-4.3}}$ & $0.0{\scriptstyle^{+0.0}_{-0.0}}$ & $48.4{\scriptstyle^{+18.8}_{-18.4}}$ & $59.3{\scriptstyle^{+18.8}_{-19.2}}$ & $23.8{\scriptstyle^{+10.3}_{-9.3}}$ \\
    \midrule
    \multirow{3}{*}{Qwen3-Coder-480B} & Few-Shot & $4.2{\scriptstyle^{+7.2}_{-4.2}}$ & $17.0{\scriptstyle^{+9.2}_{-7.1}}$ & $43.7{\scriptstyle^{+16.9}_{-18.8}}$ & $36.6{\scriptstyle^{+16.8}_{-16.6}}$ & $38.4{\scriptstyle^{+10.4}_{-10.0}}$ \\
     & Zero-Shot & $4.3{\scriptstyle^{+7.2}_{-4.3}}$ & $0.4{\scriptstyle^{+2.1}_{-0.4}}$ & $34.3{\scriptstyle^{+19.1}_{-16.5}}$ & $33.8{\scriptstyle^{+16.7}_{-14.0}}$ & $37.3{\scriptstyle^{+11.1}_{-10.4}}$ \\
     & Direct & $24.9{\scriptstyle^{+15.2}_{-13.4}}$ & $2.1{\scriptstyle^{+4.2}_{-2.1}}$ & $59.1{\scriptstyle^{+17.6}_{-17.4}}$ & $33.8{\scriptstyle^{+19.2}_{-16.9}}$ & $22.1{\scriptstyle^{+9.3}_{-8.3}}$ \\
    \bottomrule
  \end{tabular}
  \caption{%
    Results for coding LLMs across five domains.
    \textbf{SARA Numeric} and \textbf{Airline} report accuracy;
    \textbf{SARA Binary}, \textbf{USCIS-AAO}, and \textbf{Housing} report macro~F1.
    Values are reported as mean$^{+\text{upper}}_{-\text{lower}}$, where \textit{upper} and \textit{lower} denote distances from the mean to the 97.5th and 2.5th bootstrap percentiles, respectively.}
  \label{tab:coding-agent-results}
\end{table}

GPT-5.2-Codex achieves the strongest overall performance, yet its performance remains sensitive to prompts. In contrast, Qwen-based coding models frequently fail the tasks, resulting in near-zero scores in several settings. This suggests that code generation for rule-grounded reasoning is challenging for coding agents, with small prompt variations leading to catastrophic failures. Additionally, large bootstrap confidence intervals indicate substantial variance across runs, highlighting the instability of Prolog generations. Overall, these results demonstrate that, despite their strong coding capabilities, current coding agents struggle to reliably generate executable Prolog codes for complex, long-context rule reasoning tasks.

\subsection{Impact of Reasoning Effort on Performance}
\label{appendix:reasoning-effort-main}
\begin{figure*}[t]
\centering

\includegraphics[width=0.35\textwidth]{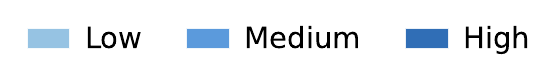}
\vspace{0.5em}

\begin{subfigure}[t]{0.85\textwidth}
\centering
\includegraphics[width=\textwidth]{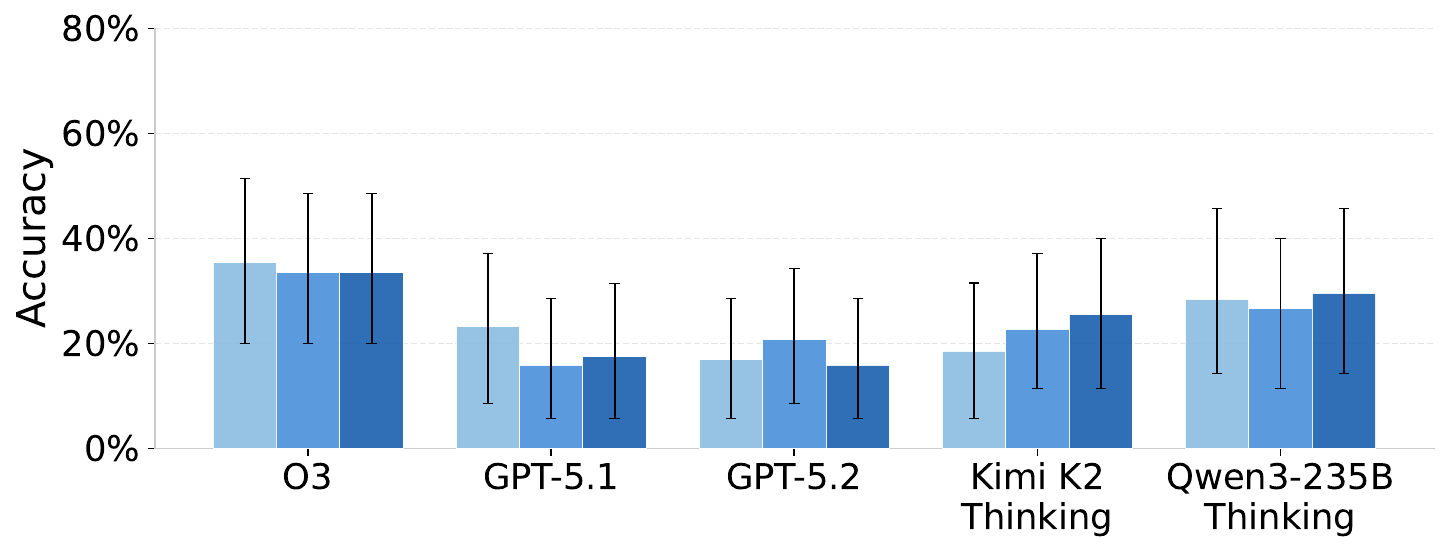}
\caption{Direct}
\end{subfigure}
\hfill
\begin{subfigure}[t]{0.85\textwidth}
\centering
\includegraphics[width=\textwidth]{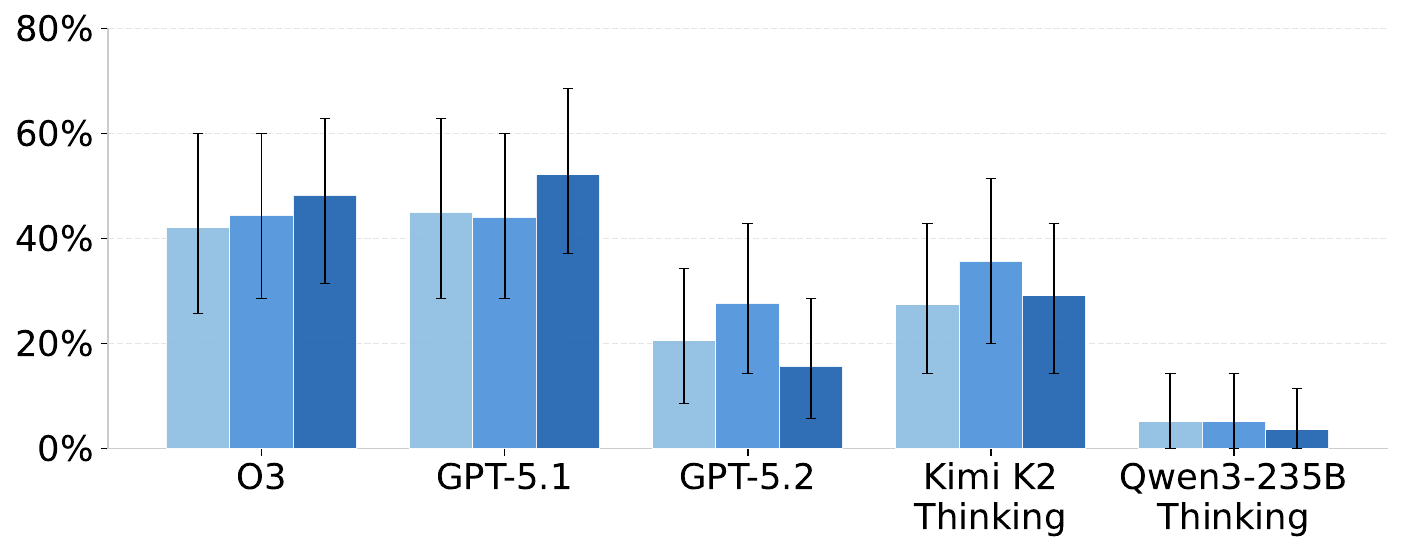}
\caption{Zero-Shot}
\end{subfigure}
\hfill
\begin{subfigure}[t]{0.85\textwidth}
\centering
\includegraphics[width=\columnwidth]{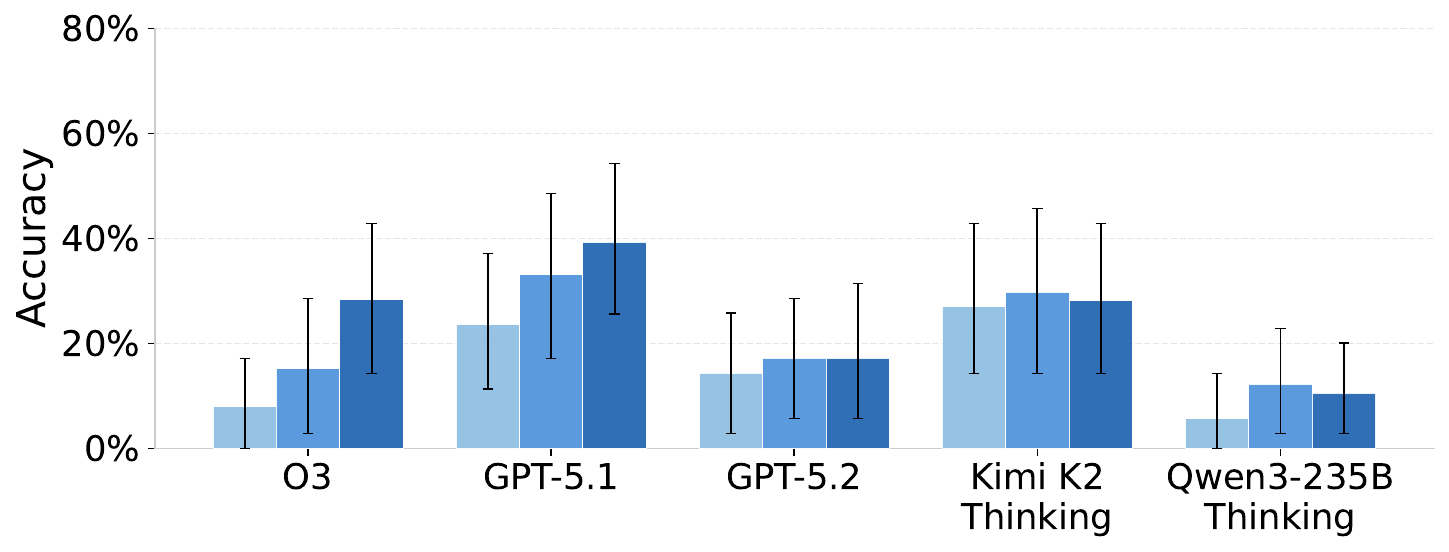}
\caption{Few-Shot}
\end{subfigure}
\hfill
\caption{Reasoning effort ablation on SARA Numeric hard cases. Each panel corresponds to a different prompting strategy (Direct, Zero-Shot, and Few-Shot). Bars indicate mean accuracy, with error bars showing 95\% bootstrap confidence intervals.}
\label{fig:reasoning-effort-sara}
\end{figure*}

We further investigate how increasing reasoning effort affects model performance on SARA Numeric. As shown in Figure \ref{fig:reasoning-effort-sara}, performance does not consistently improve with higher reasoning effort; in some cases, additional effort leads to diminishing or even negative returns.

Moreover, GPT-5.1 consistently outperforms GPT-5.2 across SARA Numeric, SARA Binary, USCIS-AAO, and Housing, and remains competitive on Airline (Table \ref{tab:main-results}). This suggests that stronger general reasoning and instruction-following capabilities do not directly translate to improved performance on \deontic. Instead, success in our setting appears to depend more critically on the ability to generate precise, executable reasoning structures.

\subsection{Additional Failure Mode Analysis}
\label{appendix:additional-error-bar}
\begin{figure*}[t]
\centering

\includegraphics[width=0.6\textwidth]{Figures/final_figures_for_paper/error_analysis_cb_exp_4/legend_error_analysis.pdf}



\begin{subfigure}[t]{\textwidth}
\centering
\includegraphics[width=\textwidth]{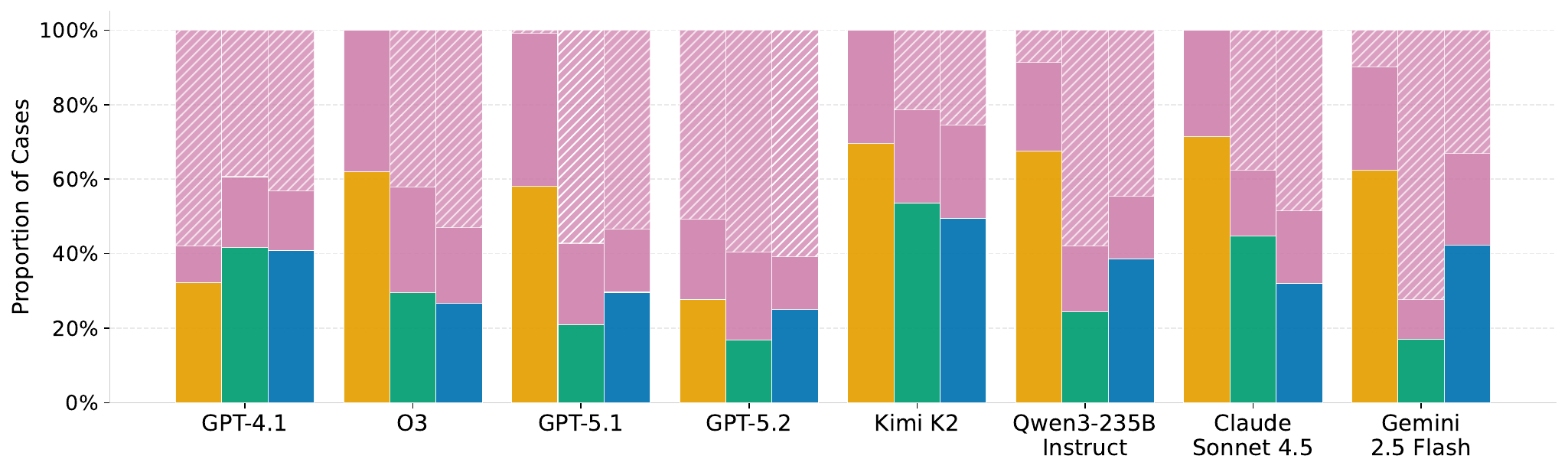}
\caption{SARA Binary}
\end{subfigure}

\vspace{2pt}

\begin{subfigure}[t]{\textwidth}
\centering
\includegraphics[width=\textwidth]{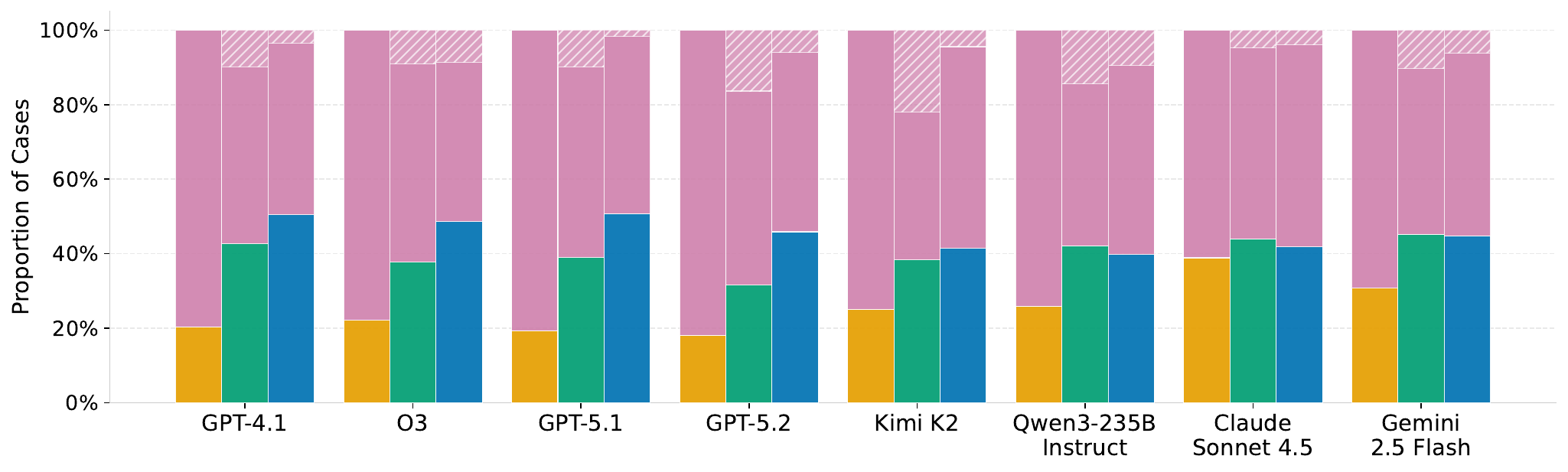}
\caption{Housing}
\end{subfigure}

\vspace{2pt}

\begin{subfigure}[t]{\textwidth}
\centering
\includegraphics[width=\textwidth]{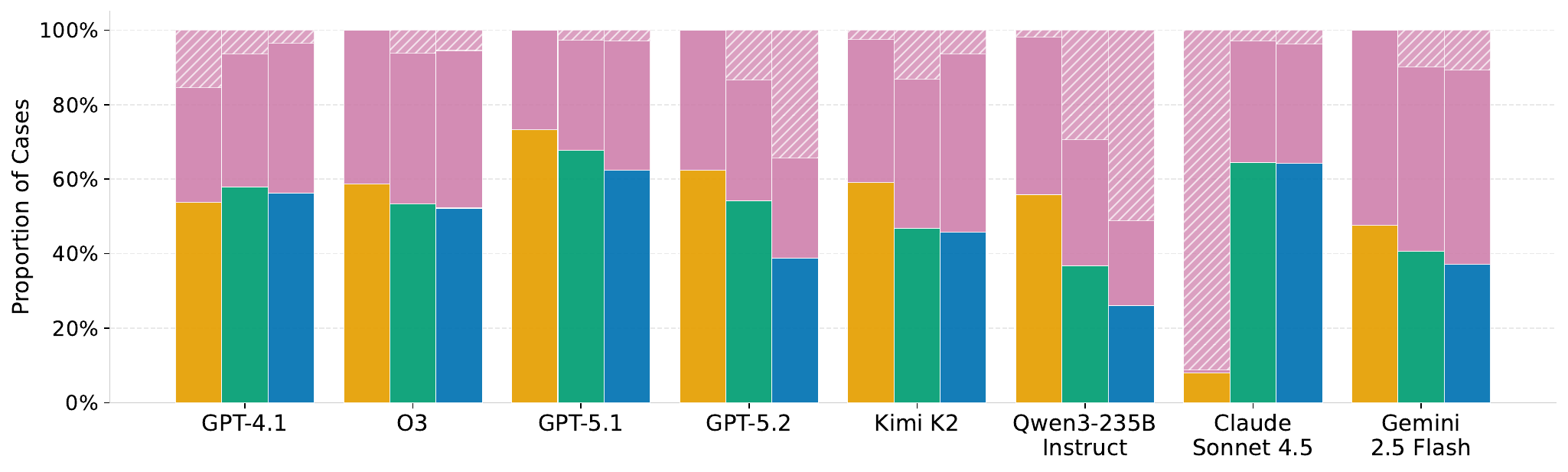}
\caption{USCIS-AAO}
\end{subfigure}

\caption{%
  Performance decomposition analysis for SARA Binary, Housing, and USCIS.
  Each model shows three bars (left to right: Direct, Zero-Shot, Few-Shot).
  The colored portion represents correct predictions;
  the solid pink portion represents wrong answers;
  the hatched pink portion represents abstentions (Prolog timeout or crash).
}
\label{fig:error-analysis-appendix}
\end{figure*}

In this section, we present additional error bar charts for SARA Binary, Housing, and USCIS-AAO in Figure \ref{fig:error-analysis-appendix}.

\section{USCIS-AAO Dataset Descriptive Statistics}
\label{appendix:uscis-dataset-stats}

Table \ref{tab:uscis-dataset-stats} presents the statistics of the USCIS-AAO dataset.
\begin{table*}[t]
\centering
\scriptsize
\setlength{\tabcolsep}{4pt}

\begin{tabular}{@{}lrrr@{}}
\toprule
Subset & Cases & Accepted & Dismissed \\
\midrule
\textbf{Total} & 242 & 121 (50.0\%) & 121 (50.0\%) \\
\quad 2022 & 11 & 2 (18.2\%) & 9 (81.8\%) \\
\quad 2023 & 19 & 17 (89.5\%) & 2 (10.5\%) \\
\quad 2024 & 188 & 78 (41.5\%) & 110 (58.5\%) \\
\quad 2025 & 24 & 24 (100.0\%) & 0 (0.0\%) \\
\addlinespace[2pt]
\textbf{Hard} & 28 & 14 (50.0\%) & 14 (50.0\%) \\
\quad 2022 & 11 & 2 (18.2\%) & 9 (81.8\%) \\
\quad 2023 & 6 & 4 (66.7\%) & 2 (33.3\%) \\
\quad 2024 & 11 & 8 (72.7\%) & 3 (27.3\%) \\
\bottomrule
\end{tabular}

\vspace{0.5em}

\begin{tabular}{@{}lrrrrrrrr@{}}
\toprule
& \multicolumn{2}{c}{Law} & \multicolumn{2}{c}{Facts narrative} & \multicolumn{2}{c}{Case Text} & \multicolumn{2}{c}{Tokens} \\
Subset & Sent. & Words & Sent. & Words & Sent. & Words & Complete case & Prolog \\
\midrule
Total & 18.8 & 300.0 & 10.7 & 320.2 & 29.5 & 620.3 & 832.5 & 884.0 \\
Hard & 16.1 & 250.9 & 10.4 & 330.3 & 26.4 & 581.2 & 778.4 & 956.0 \\
\bottomrule
\end{tabular}

\caption{Descriptive statistics for the USCIS-AAO dataset are reported in aggregate and for the hard subset. The upper panel presents overall totals, while indented subrows provide year-specific case counts and within-year percentages for Accepted and Dismissed cases. Mean token counts are computed using the \texttt{cl100k\_base} encoding.}
\label{tab:uscis-dataset-stats}
\end{table*}

\subsection{USCIS-AAO Dataset Scope and Preprocessing}
\label{appendix:uscis-dataset-process}

The AAO possesses a defined jurisdiction, function, and policy. This appendix outlines the scope of publicly available materials and details the preprocessing steps used to construct the dataset.

\paragraph{AAO Jurisdiction} The Administrative Appeals Office (AAO) conducts independent de novo appellate review of United States Citizenship and Immigration Services (USCIS) officers' decisions, applying USCIS policies and legal interpretations to each case \citep{uscis_aao_practice_manual}. The AAO functions independently from field offices and reviews all issues of fact, law, policy, and discretion. Upon appeal, the AAO re-examines the record anew, and its decision may address issues that were not previously raised or resolved.

The AAO adjudicates three primary categories of cases: appeals, motions, and certifications. Its jurisdiction is determined by both subject matter and form number, and it includes authority over motions to reopen and motions to reconsider its own decisions. Additionally, the AAO may review an initial decision at the request of USCIS officers when a case involves an unusually complex or novel issue of law or fact.

\paragraph{Dataset Temporal Scope} AAO non-precedent decisions serve as effective benchmarks because they generally reflect the application of existing law and agency policy to specific factual scenarios, rather than establishing new binding doctrine. However, the governing legal framework evolves over extended periods, as binding standards may shift due to subsequent precedent decisions, adopted AAO decisions, or revisions to USCIS policy guidance. Consequently, constructing a benchmark from cases spanning multiple years would conflate two sources of variation: differences in adjudicative reasoning under a stable legal regime and changes resulting from modifications to the legal regime itself.

Therefore, the dataset is not constructed by pooling cases from widely separated years. The dataset is assembled from a restricted subset of years (2022--2025) to ensure that the applicable legal and policy regime remains approximately stable. This approach reduces the risk of contradictions that arise not from inconsistent reasoning over similar facts, but from changes in governing rules or official guidance over time.

Let a decision at time $t$ be written as
$$D_t = E_t(F, R_t),$$

where $F$ denotes the factual record, $R_t$ denotes the set of governing rules and policy in force at time $t$, and $E_t$ is the adjudicative mapping from facts and rules to an outcome. Over extended time periods, neither $R_t$ nor $E_t$ can be assumed constant. The underlying assumption is that variation in outcomes across years is partially attributable to divergence in the governing rule sets. Restricting the benchmark to a narrower temporal window serves as an identification strategy, isolating the model's ability to model the evaluation function $E_t$ under a relatively stable regime, rather than rewarding or penalizing the model for shifts in precedent, policy, or official guidance.

\paragraph{Facts Extraction Process} The \textit{Analysis} section of each AAO case presents the office's reasoning by applying relevant legal authorities to the evidence in the record to justify the final disposition. This section typically includes evaluations of individual facts, intermediate legal conclusions, and predicate eligibility determinations. 

The primary objective of the dataset is to assess whether a model can predict the case disposition based on the facts and governing legal authorities. Accordingly, the \textit{Law} section is preserved as a near-verbatim statement of the legal authorities governing the case, including relevant statutes, regulations, precedents, and policy materials. In contrast, the \textit{Facts} value is extracted from the \textit{Analysis}, where factual statements are often interwoven with the AAO's application of law to facts. 

This separation was achieved through language-model-assisted extraction followed by human verification. The annotator received only the factual narrative and was instructed to label and exclude any case containing text that revealed the office's application of legal authorities to the facts, an intermediate eligibility determination, or the ultimate outcome.  In accordance with this protocol, the released dataset underwent a secondary review, which identified additional errors, including text truncation and further contamination. The \textit{facts} narrative was subsequently re-generated and partially rewritten to ensure that it does not explicitly disclose the case's final resolution.

This procedure has important limitations. First, the boundary between factual narrative and legal analysis is partly determined by the annotation protocol and is not entirely objective. Second, the quality of the final artifact depends on the accuracy of the extraction model and the consistency of the human review process. Implementing a stricter annotation rubric, involving multiple annotators, and adjudicating disagreements would further enhance the reliability of this filtering procedure.

\section{Dataset Examples}
\label{appendix:examples}

We present one representative instance from each domain in \textsc{DeonticBench}.
For each domain, we show the relevant statutes (abbreviated where lengthy), the case
description, the question, the ground-truth label, and the reference Prolog program
(abbreviated where lengthy; omissions marked with \texttt{\% \ldots}).

\lstset{
  basicstyle=\small\ttfamily,
  upquote=true,
  showstringspaces=false,
  breaklines=true,
  columns=fullflexible,
  keepspaces=true,
  frame=single,
  backgroundcolor=\color{gray!10},
  rulecolor=\color{gray!55},
  xleftmargin=0.5em,
  xrightmargin=0.5em,
  aboveskip=4pt,
  belowskip=4pt
}

\newtcolorbox{examplecard}[1]{
  breakable,
  colback=gray!10,
  colframe=gray!55,
  colbacktitle=gray!60,
  coltitle=white,
  fonttitle=\small\bfseries\sffamily,
  title={#1},
  left=6pt,
  right=6pt,
  top=6pt,
  bottom=6pt,
  boxrule=0.6pt,
  arc=3pt
}

\newcommand{\exampledivider}{%
  \par\medskip
  \noindent\textcolor{gray!60}{\rule{\linewidth}{0.4pt}}%
  \par\medskip
}

\subsection{SARA Numeric: U.S.\ Tax Reasoning}
\label{app:examples:sara}

\begin{examplecard}{SARA Numeric --- Example Instance}
\ttfamily
\textbf{Statutes (excerpt)}\par
The SARA domain encodes a subset of the U.S. Internal Revenue Code, including
\S\S\,1 (tax rates), 2 (filing status), 63 (taxable income), 68 (itemized deduction
limitation), 151--152 (exemptions), 3301/3306 (FUTA), and 7703 (marital status).\par

\vspace{3pt}
\textbf{Representative fragment}\par
\textbf{\S\,1(a). Married individuals filing joint returns and surviving spouses.}\par
A tax is imposed on the taxable income of every married individual who makes a single return
jointly with his spouse, and every surviving spouse, as follows:\par
(i) 15\% if not over \$36,900;\par
(ii) \$5,535 plus 28\% of excess over \$36,900 if over \$36,900 but not over \$89,150;\par
\ldots\par
(v) \$75,528.50 plus 39.6\% of excess over \$250,000.\par

\exampledivider
\textbf{Case}\par
Alice and Harold got married on Sep 3rd, 1992. Harold and Alice have a son, born Jan 25th,
2000. Harold died on Feb 28th, 2016. They had been living in the same house since 1993,
maintained by Alice. Alice and her son continued doing so after Harold's death. Alice's
gross income for the year 2017 was \$236,422. Alice employed Bob, Cameron, Dan, Emily,
Fred, and George for agricultural labor from Sep 9th to Oct 1st 2017, paying them
\$5,012 each. Alice takes the standard deduction in 2017.\par

\exampledivider
\textbf{Question}\par
How much tax does Alice have to pay in 2017?\par

\exampledivider
\textbf{Label}\par
\$68,844
\end{examplecard}

\noindent\textbf{Reference Prolog (abridged).}\par
\begin{lstlisting}[language=Prolog]
% Facts from the case
spouse('Alice','Harold').
died('Harold',2016).
not_remarried('Alice',2017).
joint_return_possible('Alice','Harold',2016).
child('Alice','Son').
birth_year('Son',2000).
principal_abode('Son','Alice',2017).
not_joint_return('Son',2017).
maintains_home('Alice',2017).
half_cost('Alice',2017).
gross_income('Alice',2017,236422).
takes_standard_deduction('Alice',2017).
wage('Alice','Bob',2017,5012,agricultural,cash).
% ... (wages for Cameron, Dan, Emily, Fred, George) ...

% Filing status: surviving spouse check
surviving_spouse(P,Year) :-
    spouse(P,Sp), died(Sp,DY),
    (DY is Year-1 ; DY is Year-2),
    not_remarried(P,Year), maintains_home(P,Year), half_cost(P,Year),
    dependent(P,_,Year), joint_return_possible(P,Sp,DY).

filing_status(P,Year,surviving_spouse) :- surviving_spouse(P,Year).
filing_status(P,Year,single) :-
    \+ married(P,Year), \+ surviving_spouse(P,Year), \+ head_of_household(P,Year).

% Tax brackets for surviving spouse (Sec. 1(a))
tax_from_brackets(surviving_spouse,TI,Tax) :-
    (TI =< 36900  -> Tax is 0.15*TI ;
     TI =< 89150  -> Tax is 5535    + 0.28*(TI-36900)  ;
     TI =< 140000 -> Tax is 20165   + 0.31*(TI-89150)  ;
     TI =< 250000 -> Tax is 35928.50 + 0.36*(TI-140000) ;
     Tax is 75528.50 + 0.396*(TI-250000)).

% Employer agricultural tax (FUTA Secs. 3301/3306)
employer_tax(P,Year,Tax) :-
    wages_for_employment(P,Year,Total), Tax is 0.06*Total.

% Total tax
tax(P,Year,Tax) :-
    income_tax(P,Year,IT),
    (employer_tax(P,Year,ET) -> true ; ET=0),
    Sum is IT + ET, round2(Sum,Tax).

:- tax('Alice', 2017, Tax), format('Tax result: ~w~n', [Tax]).
:- halt.
\end{lstlisting}

\subsection{SARA Binary: U.S.\ Tax Entailment Checking}
\label{app:examples:sara-binary}

\begin{examplecard}{SARA Binary --- Example Instance}
\ttfamily
\textbf{Note on format}\par
Unlike SARA Numeric, SARA Binary is an entailment task: given a case description and a
statement about the applicable law, the model must determine whether the statement is
\emph{entailed} or \emph{contradicted} by the case under the relevant statutory rules.\par

\vspace{3pt}
\textbf{Statutes (excerpt)}\par
The SARA Binary domain draws on the same U.S.\ Internal Revenue Code provisions as SARA
Numeric, focusing on Sections 151--152 (exemptions and dependents).\par

\vspace{3pt}
\textbf{Relevant fragment}\par
\textbf{Sec.\,152(c)(1). Qualifying child.}
An individual is a qualifying child of a taxpayer for a taxable year if the individual
bears a specified relationship to the taxpayer (Sec.\,152(c)(2)), has the same principal
place of abode as the taxpayer for more than one-half of the year (Sec.\,152(c)(3)), meets
the age requirement, has not provided more than one-half of his own support, and has not
filed a joint return.\par
\textbf{Sec.\,152(b)(1).}
A qualifying child of a taxpayer under Sec.\,152(c)(1) is treated as a dependent of that
taxpayer; the provision ``applies'' to an individual for any taxable year beginning in the
calendar year in which they are a qualifying child of any taxpayer.\par

\exampledivider
\textbf{Case}\par
Alice has a son, Bob, who satisfies section 152(c)(1) for the year 2015. Bob has a son,
Charlie, who satisfies section 152(c)(1) for the year 2015. Alice's income in 2015 was
\$504,598. Bob had no income in 2015.\par

\exampledivider
\textbf{Question ('Statement to verify')}\par
Section 152(b)(1) applies to Bob for the year 2015.\par

\exampledivider
\textbf{Label}\par
Entailment
\end{examplecard}

\noindent\textbf{Reference Prolog.}\par
\begin{lstlisting}[language=Prolog]
/* ---------- Statutory rules ---------- */

/* Sec. 152(c)(1): qualifying child => dependent
   (subject to Sec. 152(b)(2) married-joint-return exception) */
dependent(Child, Taxpayer, Year) :-
    qualifying_child(Child, Taxpayer, Year),
    \+ married_joint_return(Child, Year).

/* Sec. 152(b)(1): applies to an individual for a year
   whenever the individual is a dependent of some taxpayer
   for a taxable year beginning in that calendar year.       */
section_152b1_applies(Individual, Year) :-
    dependent(Individual, _Someone, Year).

/* ---------- Facts from the case ---------- */
qualifying_child(bob, alice, 2015).
/* Bob satisfies Sec. 152(c)(1) for Alice; no joint-return fact asserted */

/* ---------- Verification query ---------- */
:- (  section_152b1_applies(bob, 2015)
   -> format('Result: Entailment~n')
   ;  format('Result: Contradiction~n')
   ).
:- halt.
\end{lstlisting}

\subsection{Airline: Baggage Policy Reasoning}
\label{app:examples:airline}

\begin{examplecard}{Airline --- Example Instance}
\ttfamily
\textbf{Statutes (excerpt)}\par
The airline domain encodes American Airlines' published baggage policies covering carry-on
allowances, checked-bag fees by route and cabin class, and special-item surcharges.\par

\vspace{3pt}
\textbf{Policy fragment}\par
\textbf{Carry-on.} 1 carry-on (max 22\,$\times$\,14\,$\times$\,9 in) and 1 personal item
(max 18\,$\times$\,14\,$\times$\,8 in) allowed in all cabins.\par
\textbf{First checked bag (domestic U.S./PR/USVI).}
Basic Economy and Main Cabin: \$40. Premium Economy, Business, First: \$0.\par
\textbf{Overweight fee.} Bags 51--70\,lbs: \$100 surcharge.
Bags 71--100\,lbs: \$200 surcharge.\par

\exampledivider
\textbf{Case}\par
Linda is a Business Class passenger flying from Charlotte to Phoenix with the following items:
(1) a backpack: 18\,$\times$\,13\,$\times$\,6 in, 8\,lbs;
(2) a luggage box: 41\,$\times$\,20\,$\times$\,16 in, 95\,lbs;
(3) a backpack: 38\,$\times$\,24\,$\times$\,18 in, 74\,lbs;
(4) a backpack: 37\,$\times$\,16\,$\times$\,10 in, 54\,lbs;
(5) a backpack: 43\,$\times$\,25\,$\times$\,20 in, 52\,lbs.
Linda's flight ticket is \$186.\par

\exampledivider
\textbf{Question}\par
What is the total cost (including flight ticket, checked bag fees, and special-needs costs)
according to the policies?\par

\exampledivider
\textbf{Label}\par
\$1,166
\end{examplecard}

\noindent\textbf{Reference Prolog (abridged).}\par
\begin{lstlisting}[language=Prolog]
flight_cost(186).
% checked bags: bag(Id,Length,Width,Height,Weight)
bag(2,41,20,16,95).  bag(3,38,24,18,74).
bag(4,37,16,10,54).  bag(5,43,25,20,52).
checked_bag_ids([2,3,4,5]).

% overweight fee (charged bags have 50-lb threshold)
overweight_fee(W,Thresh,Fee) :-
    (W =< Thresh -> Fee=0 ; W=<53 -> Fee=30 ;
     W=<70 -> Fee=100 ; W=<100 -> Fee=200).

% oversize fee (domestic)
oversize_fee(Sum,Fee) :-
    (Sum =< 62 -> Fee=0 ; Sum =< 65 -> Fee=30 ; Fee=200).

penalty(BagId,charged,Pen) :-
    bag(BagId,L,W,H,Weight),
    dims_sum(L,W,H,Sum),
    oversize_fee(Sum,OS), overweight_fee(Weight,50,OW),
    Pen is max(OS,OW).

% Business Class: first 2 bags free; 3rd $150; 4th+ $200
bag_fee(1,0). bag_fee(2,0). bag_fee(3,150).
bag_fee(N,200) :- N>=4.

% Minimise total cost over all choices of 2 complimentary bags
total_cost(MinTotal) :-
    checked_bag_ids(L),
    findall(Total, (choose_two(L,F1,F2), cost_for_choice(F1,F2,Total)), Totals),
    min_list(Totals,MinTotal).

:- total_cost(Total), format('Total cost: ~w~n', [Total]).
:- halt.
\end{lstlisting}

\subsection{USCIS-AAO: Immigration Appeals}
\label{app:examples:uscis}

\begin{examplecard}{USCIS-AAO --- Example Instance}
\ttfamily
\textbf{Statutes (excerpt)}\par
A delivery bond creates a contract between the U.S.\ Government and an obligor.
A breach occurs upon substantial violation of a bond's conditions (8~C.F.R.\ \S\,103.6(e));
conversely, substantial performance releases the obligor from liability (8~C.F.R.\ \S\,103.6(c)(3)).
Several factors inform whether a violation is substantial: the extent of the violation;
whether it was intentional or accidental; whether it was in good faith; and whether the
obligor took steps to comply.
\textit{Matter of Kubacki}, 18 I\&N Dec.\ 43, 44 (Reg'l Comm'r 1981).\par

\exampledivider
\textbf{Case}\par
The Obligor posted an ICE Form~I-352 immigration delivery bond as security for a bonded
foreign national.
ICE declared the bond breached after the bonded foreign national was not produced in
response to a written request; the Obligor appealed seeking reinstatement.
On appeal, the Obligor argued it could not produce the foreign national because the
Notice to Appear (Form~I-862) lacked a time and place for the hearing, and that under
\textit{Pereira v.\ Sessions} the defective NTA rendered the proceedings invalid, thereby
negating any substantial violation of the bond's terms.\par

\exampledivider
\textbf{Question}\par
Should this case be accepted or dismissed?\par

\exampledivider
\textbf{Label}\par
Dismissed
\end{examplecard}

\noindent\textbf{Reference Prolog (abridged).}\par
\begin{lstlisting}[language=Prolog]
% Facts
delivery_bond_posted. ice_declared_bond_breached.
written_request_to_produce_issued. bonded_foreign_national_not_produced.
obligor_appealed. nta_lacked_time_or_place.
obligor_cites_pereira. obligor_claims_nta_invalid.

% Core delivery-bond condition
condition_to_produce_on_written_request.

violation_occurred :-
    condition_to_produce_on_written_request,
    written_request_to_produce_issued,
    bonded_foreign_national_not_produced.

% Kubacki substantiality factors
extent_of_violation_significant :- bonded_foreign_national_not_produced.
% record does not establish good faith, steps to comply, or accidental nonproduction

% NTA defect (Pereira) is not a defence to nonproduction under the bond
nta_defect_not_a_defense_to_production_obligation :-
    nta_lacked_time_or_place,
    obligor_cites_pereira,
    obligor_claims_nta_invalid.

substantial_violation :-
    violation_occurred,
    extent_of_violation_significant,
    nta_defect_not_a_defense_to_production_obligation.

substantial_performance :- \+ bonded_foreign_national_not_produced.

eligibility_met :-
    substantial_performance, \+ substantial_violation.

decision(Result) :-
    (eligibility_met -> Result = 'Accepted' ; Result = 'Dismissed').
\end{lstlisting}

\subsection{Housing: Eviction Law Q\&A}
\label{app:examples:housing}

\begin{examplecard}{Housing --- Example Instance}
\ttfamily
\textbf{Note on format}\par
Unlike the other three domains, Housing instances contain no narrative case
description---each instance is a (state, statutes, question) triple, requiring the model to reason over the provided
statute excerpts to answer a binary yes/no question.\par

\vspace{3pt}
\textbf{Statutes}\par
\textbf{MICH.\ COMP.\ LAWS \S\,600.5704.}
The district court, municipal courts and the common pleas court of Detroit have jurisdiction
over summary proceedings to recover possession of premises under this chapter.\par

\textbf{MICH.\ COMP.\ LAWS \S\,600.5706(3).}
In districts where the district court is not operative, the municipal court of the city in
which the premises or any part of the premises are situated is a proper court in which to
commence and try summary proceedings.\par

\exampledivider
\textbf{State}\par
Michigan\par

\exampledivider
\textbf{Question}\par
Are eviction cases first heard in municipal court?\par

\exampledivider
\textbf{Label}\par
Yes
\end{examplecard}

\noindent\textbf{Reference Prolog (abridged).}\par
\begin{lstlisting}[language=Prolog]
state(michigan).

statute(mich_comp_laws_600_5704).
statute_of_state(mich_comp_laws_600_5704, michigan).
jurisdiction(mich_comp_laws_600_5704, district_court).
jurisdiction(mich_comp_laws_600_5704, municipal_court).
refers_to(mich_comp_laws_600_5704, summary_proceedings).
% ...

statute(mich_comp_laws_600_5706).
statute_of_state(mich_comp_laws_600_5706, michigan).
jurisdiction(mich_comp_laws_600_5706, district_court).
jurisdiction(mich_comp_laws_600_5706, municipal_court).
refers_to(mich_comp_laws_600_5706, summary_proceedings).
% ...

eviction_cases_first_heard_in_municipal_court :-
    statute_of_state(Law, _State),
    jurisdiction(Law, municipal_court),
    refers_to(Law, summary_proceedings).

% ...

housing_answer(Result) :-
    derived_answer(Result).

main :-
    housing_answer(Result),
    format('housing_answer(~w).~n', [Result]).
:- initialization(main, main).
\end{lstlisting}

\end{document}